\documentclass{article}
\usepackage{setup/iclr2027_conference}
\iclrpreprintcopy   % \iclrfinalcopy for the camera-ready header; neither for the anonymous submission

\usepackage{lipsum}
\usepackage[T1]{fontenc}
\usepackage{amsmath, amssymb, amsthm, amsfonts, mathtools, mathrsfs}
\usepackage{graphicx, float, subcaption}
\usepackage{wrapfig}
\usepackage{booktabs, tabularx, colortbl}
\usepackage[dvipsnames]{xcolor}
\usepackage{hyperref, cleveref, url}
\usepackage{times, microtype, pifont, tikz}
\usepackage{inconsolata}
\usepackage[most]{tcolorbox}
\tcbuselibrary{listings}

\theoremstyle{definition}
\newtheorem{definition}{Definition}

\NewDocumentCommand{\alexi}
{ mO{} }{\textcolor{red}{\textsuperscript{\textit{Alexi}}\textsf{\textbf{\small[#1]}}}}
\hypersetup{
    colorlinks=true,
    linkcolor=black,
    citecolor=black,
    urlcolor=blue,
}
\setcitestyle{authoryear,open={(},close={)}}

\nocite{*}

\theoremstyle{definition}

\definecolor{checkgreen}{HTML}{388C3C}
\definecolor{xred}{HTML}{CE5E5E}
\definecolor{batlowBlue}{HTML}{103D5F}
\definecolor{domDem}{HTML}{3F6F8E}     % first input domain, steel blue sampled from the Table 2 illustrations
\definecolor{domBank}{HTML}{765030}    % second input domain, stall-wood brown from the market colony illustration
\definecolor{newConcept}{HTML}{22394F} % the invented entity
\definecolor{emergentCol}{HTML}{B74D34} % emergent structure, the awning red of the market colony illustration

\newtcblisting{promptbox}[1][Prompt]{%
  enhanced, breakable, arc=2pt, boxrule=0.6pt,
  colback=batlowBlue!4, colframe=batlowBlue!72!black,
  coltitle=white, fonttitle=\bfseries\small, title={#1},
  left=4pt, right=4pt, top=2pt, bottom=2pt, listing only,
  listing options={basicstyle=\fontsize{7.5}{9}\selectfont\ttfamily, breaklines=true,
    columns=fullflexible, keepspaces=true, breakindent=0pt, breakautoindent=false}%
}
\newtcolorbox[auto counter, crefname={task}{tasks}, Crefname={Task}{Tasks}]{taskbox}[2][]{
    enhanced, breakable,
    colback=batlowBlue!4!white, colframe=batlowBlue!72!black,
    colbacktitle=batlowBlue!72!black, coltitle=white, fonttitle=\bfseries\small,
    boxrule=0.6pt, arc=1.2mm,
    left=2.5mm, right=2.5mm, top=1.5mm, bottom=1.5mm, toptitle=0.6mm, bottomtitle=0.6mm,
    before skip=6pt, after skip=6pt,
    fontupper=\small, title={#2 Task}, label={#1},   % unnumbered: 'Association Task' etc. (author, 2026-09-24)
}

\NewDocumentCommand{\heng}
{ mO{} }{\textcolor{red}{\textsuperscript{\textit{Heng}}\textsf{\textbf{\small[#1]}}}}

\newtcolorbox{researchquestion}[1]{
    enhanced,
    breakable,
    colback=RoyalBlue!5!white,
    colframe=RoyalBlue!70!black,
    colbacktitle=RoyalBlue!70!black,
    coltitle=white,
    fonttitle=\bfseries,
    boxrule=0.6pt,
    arc=1.2mm,
    left=3mm, right=3mm, top=2.5mm, bottom=2.5mm,
    title={#1},
}

\title{Measuring Combinatorial Creativity in \\ Large Language Models and Humans}
\author{Samuel Schapiro\textsuperscript{1,3 $\dagger$}
\And
Akshay K. Jagadish\textsuperscript{2}
\And
Alexi Gladstone\textsuperscript{6}
\And
Qiawen Ella Liu\textsuperscript{8}
\And 
Sumuk Shashidhar\textsuperscript{6}
\And
Jonah Black\textsuperscript{6}
\And
Royce Moon\textsuperscript{3, 9}
\And
Dilek Hakkani-Tur\textsuperscript{6}
\And
Heng Ji\textsuperscript{6}
\And
Claire E. Stevenson\textsuperscript{7}
\And
Lav R. Varshney\textsuperscript{3}
\And
Roger E. Beaty\textsuperscript{5}
\And
Thomas L. Griffiths\textsuperscript{2, 8}
\And
James A. Evans\textsuperscript{1, 4}
\AND
{}\\[-15pt]
\normalfont\normalsize
\begin{tabular}{l}
\textsuperscript{1}Knowledge Lab, University of Chicago \quad
\textsuperscript{2}Princeton AI Lab\\
\textsuperscript{3}AI Innovation Institute, Stony Brook University\quad
\textsuperscript{4}Santa Fe Institute \\
\textsuperscript{5}Department of Psychology, Pennsylvania State University \\
\textsuperscript{6}School of Computer Science, University of Illinois Urbana-Champaign \\
\textsuperscript{7}Department of Psychology, University of Amsterdam \\
\textsuperscript{8}Department of Psychology, Princeton University \quad
\textsuperscript{9}University of Michigan \\ 
\end{tabular}
}

\begin{document}
\maketitle

\vspace{-25pt}
\begin{figure}[H]
    \centering
    \includegraphics[width=0.7\linewidth]{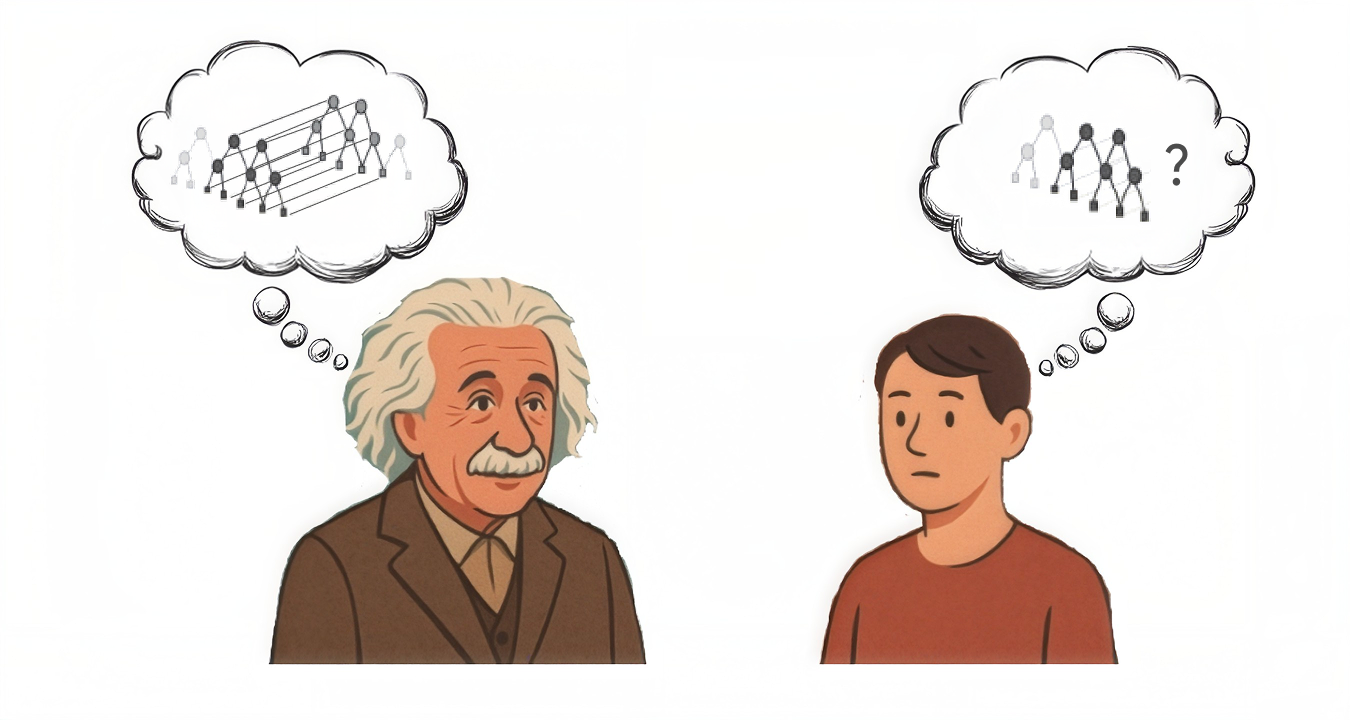}
    \caption{ \textbf{Overview.} Combinatorial creativity, spanning association, analogy, and blending, describes the process of combining familiar ideas in unfamiliar ways, and is recognized as central to scientific and technological progress \citep{creative_comb_rep, boden, surprising_combinations, simonton_discovery_invention_as_combinatorial}.}
    \label{fig:overview}
\end{figure}
{\renewcommand{\thefootnote}{\fnsymbol{footnote}}\footnotetext[2]{Part of this work done while visiting the Simons Institute for the Theory of Computing, UC Berkeley.}}
\begin{abstract} 
Combinatorial creativity, the process of combining familiar ideas in unfamiliar ways, is one of the foremost engines of scientific, technological, and artistic progress. 
%While large language models (LLMs) can fluently produce novel text relative to their training data, their capacity to combine information in novel \emph{and} useful ways, in order to invent new concepts, remains unclear. 
This work introduces \textbf{\textsc{Kombine}}, an LLM benchmark for combinatorial creativity tasking models to discover surprising, original, and useful associations, analogies, and blends between unrelated entities, and to use those analogies and blends to \emph{invent novel concepts by re-combining the parts of existing ones}. In a large-scale study across 35 LLMs, we find that: \textbf{(1)} LLMs are 10--16 percentage points better at discovering associations and analogies than blends, and \textbf{(2)} exhibit an \emph{abstraction bottleneck} on blending, in which even frontier LLMs struggle to find a shared abstraction between inputs and instantiate a coherent blended space. \textbf{(3)} LLMs also exhibit ``inventive homogeneity,'' in which one in five concepts invented for a given item reuses properties invented by another model. Moreover, \textbf{(4)} models are $14\times$ more likely to reuse other invented properties on blends than analogies, and \textbf{(5)} models from the same provider are $2.3\times$ more likely to reuse properties than those from different providers. Next, in a comparative study involving 75 human participants on the same tasks, we find that \textbf{(6)} humans score below nearly every LLM on each task and that \textbf{(7)} both humans and LLMs find blending most difficult, but association is easier than analogy for LLMs, while analogy is easier than association for humans. Lastly, \textbf{(8)} humans and LLMs access distinct regions of semantic space on each task, suggesting that combinatorial creativity could benefit from systematic human-AI collaboration to enhance conceptual and semantic diversity. In summary, we provide a framework that can be used to assess progress towards LLMs, AI systems, and human-AI collaborations more broadly, that can elicit combinatorial creativity in a setting relevant to scientific discovery and technological innovation.
\end{abstract}

\section{Introduction} \label{sec:introduction}
Einstein famously remarked that ``Combinatory play seems to be the essential feature in productive thought,'' \citep[p. 142]{hadamard} referring to the cognitive processes he believed underpinned creative insight in mathematics and the sciences. Indeed, there is a rich body of psychological literature modeling creativity as a combinatorial process over representations and abstractions \citep{koestler_creation,creativity_in_science,simonton_discovery_invention_as_combinatorial, simonton_bvsr_2010, creative_comb_rep}, and a growing body of work in the science of science showing that highly impactful scholarly work tends to comprise more distant combinations of ideas, methods, and concepts \citep{surprising_combinations, fang2025generalization}. This form of creativity elicited by making unfamiliar combinations of familiar ideas has been defined by \citet{boden} as \emph{combinatorial (or combinational) creativity}, and in recent years interest in evaluating and enhancing this ability among LLMs, especially for scientific applications, has steadily grown \citep{roll_the_dice, gu2025llmsrealizecombinatorialcreativity, artiles2026alienspacesciencesampling, shen2026unlockingllmcreativityscience, spark, jagadish2026closing}.

In the cognitive science literature, three main forms of combinatorial creativity are studied, as pictured in \Cref{fig:cc_overview}. First, \textbf{associative creativity} describes the ability to make distant connections between otherwise unrelated ideas \citep{associative_basis}, and has been studied extensively in both humans and LLMs \citep{schapiro2026assessingcreativitylargelanguage} through tests like the Divergent Association Task (DAT) \citep{Olson2021NamingCreativity}, its variants like the Conditional DAT \citep{nakajima2026divergentcreativityhumanbasedevaluation} and the Remote Associates Test (RAT) \citep{associative_basis}, and a recent associative creativity benchmark by \citet{wadhwa2026createtestingllmsassociative}. Second, \textbf{analogy} involves using knowledge from a familiar domain to infer or generate knowledge about a different, but somehow connected domain \citep{gentner1983, petersen2026modelling, liu2026large, holyoak2025humanedge}. Recent work has explored using analogical reasoning to improve creativity in LLMs for scientific discovery \citep{shen2026unlockingllmcreativityscience} and technological innovation \citep{liu2026assessingeffectcrossdomainmapping}. 
Lastly, conceptual \textbf{blending} \citep{the_way_we_think, concept_blending_comp_fr} involves selectively projecting structure from two input spaces into a new blended space with emergent properties of its own, an ability implicated in the formation of new concepts \citep{thagard1984conceptual, Marchiori2024WhatDisruption}. 

In this work, we introduce \textbf{\textsc{Kombine}}, a combinatorial creativity benchmark tasking models to \emph{discover} associations, analogies, and blends between random entity pairs with no known relationship (e.g., \texttt{a blue whale} and \texttt{a mattress}). Then, using these discovered relationships, models are tasked to invent novel concepts by re-combining the parts of existing ones, employing the structure mappings afforded by analogies and blends. In a large-scale study across 35 LLMs, including the latest frontier systems such as GPT-6 Astra and Claude Fable 5.1, we make five key findings: \textbf{(1)} Models are, on average, 10--16 percentage points better at discovering associations and analogies than blends. \textbf{(2)} Models exhibit an \emph{abstraction bottleneck} on blending, in which they struggle to find shared abstractions between inputs in order to instantiate a coherent blended space. \textbf{(3)} Models also exhibit \emph{inventive homogeneity}, in which one in five concepts invented for a given item reuses two or more properties invented by another model. \textbf{(4)} Further analysis reveals that inventive homogeneity is more pronounced among models from the same provider, which are $2.3\times$ more likely to reuse properties than those from different providers. \textbf{(5)} Moreover, models are $14\times$ more likely to reuse other invented properties on blends than analogies.

Next, in a comparative study recruiting 75 human participants on the same tasks, we find that: \textbf{(6)} Humans struggle to find valid associations, analogies, and blends between unrelated entities, scoring below nearly every LLM on each task. \textbf{(7)} Both humans and LLMs find blending most difficult, but association is easier than analogy for LLMs, while analogy is easier than association for humans. Finally, \textbf{(8)} humans and LLMs access distinct regions of semantic space on each task, suggesting that combinatorial creativity could benefit from systematic human-AI collaboration to enhance conceptual and semantic diversity.

In summary, \textsc{Kombine} offers an open-ended framework for tracking progress towards LLMs, AI systems, and human-AI interactions more broadly, that can elicit combinatorial creativity in a setting relevant to scientific discovery and technological innovation.

\begin{figure}[t]
\centering
\renewcommand{\arraystretch}{1.25}
\captionof{table}{\textbf{Forms of combinatorial creativity assessed in \textsc{Kombine}.} Each task assesses an ability recognized in cognitive science as integral to combinatorial creativity \citep{boden, bvsr_update}}
\label{tab:forms}
{\small
\begin{tabular}{@{}p{2cm} p{7cm} p{4.4cm}@{}}
\toprule
\textbf{Task (\S \ref{sec:benchmark})} & \textbf{Creativity Mechanism(s) \& Source(s)} & \textbf{Example} \\
\midrule
Association & Remote association \citep{associative_basis} & \emph{Einstein} $\to$ \emph{violin} $\to$ \emph{Mozart} \\
Analogy & Analogy \citep{gentner1983,rumelhart1973model} and metaphor \citep{lakoff1980} & \emph{immune system} $:$ \emph{body} $::$ \newline \emph{national defense} $:$ \emph{nation} \\
Blending & Conceptual blending \citep{the_way_we_think}; conceptual combination \citep{thagard1984conceptual} & \emph{organism} $+$ \emph{machine} $=$ \emph{cyborg} \\
\bottomrule
\end{tabular}
}
\input{media/fig_02_cc_overview}
\end{figure}

\section{Background and Related Work} \label{sec:background}
Before introducing our benchmark, we start by providing background on creativity, describing how creativity relates to forms of generalization, and discussing related work.

\textbf{Creativity and its Types} \quad Though creativity can refer to a person, process, product, or press (environment) \citep{4ps}, in the study of computationally creative systems, it is most common to adopt the product or process view \citep{limit_theorems, green2024process}, as we do here. In this case, the creative process involves \emph{the generation of novel, useful, and surprising products} \citep{simonton_bvsr_2010, simonton_discovery_invention_as_combinatorial, boden, limit_theorems, trans_graphical_theory}. \citet{boden} famously distinguishes between three types of creative processes: combinatorial, exploratory, and transformational creativity. The first models the creation of new artifacts as combinations of existing elements in a space of possible components, and is the focus of the present study. Consider recipe design \citep{chef_watson, mizrahi2026‍}, for example, where new recipes are generated by taking combinations of existing ingredients in varying proportions. Detailed descriptions of exploratory, transformational creativity, and other types of creativity (e.g., divergent, convergent thinking), are deferred to \Cref{app:additional_creativity_background}. Next, we detail various forms of combinatorial creativity recognized in psychological literature.

\textbf{Combinatorial Creativity} \quad Although \citet{boden} was the first to formally define combinatorial (or combinational) creativity as \emph{the combination of familiar ideas in unfamiliar ways}, the study of this ability traces back to at least \citet{hadamard}, who provided a survey of introspective accounts from famous mathematicians, scientists, and even musical composers, in which creative ideation was described as a combinatorial process. \citet{associative_basis} argued that creative thinking involved associating or combining mental representations, with more distant associations correlated with more creative artifacts, and designed the Remote Association Test (RAT) to assess this ability.
\footnote{This test, however, has come under extensive scrutiny \citep{bvsr_update, lee2014measure} due to operationalizing associative creativity via a convergent thinking instrument. As Simonton notes, ``the test should have been open-ended, with no pre-determined correct answers – like asking participants to generate their own remote associates'' (as our benchmark in \Cref{sec:benchmark} properly addresses).}
Since then, empirical studies have found associative capacity to underpin human creativity, especially in scientific domains \citep{creativity_in_science, beaty2023associative, benedek2023role, creative_comb_rep, simonton_discovery_invention_as_combinatorial, koestler_creation}. Hereafter, we refer to this ability as \textbf{associative creativity}. Subsequent work provided more structured accounts of how information can be combined across concepts and domains---such as through analogy \citep{rumelhart1973model,petersen2026modelling, liu2026large}, bisociation \citep{koestler_creation}, or blending \citep{the_way_we_think}---emphasizing the pivotal role of cross-domain mappings in scientific discovery and technological innovation \citep{gentner_kepler, Thagard1988ComputationalScience, holyoak_thagard1995}. In particular, Gentner's \citeyearpar{gentner1983} structure mapping theory formalized \textbf{analogical creativity} as systematic, structure-preserving alignment between concepts (e.g., \texttt{nucleus} is to \texttt{atom} as \texttt{sun} is to \texttt{solar system}) in ways which enable one to project structure from a source domain to infer or generate knowledge about a target domain. Similarly, \textbf{blending creativity} involves cross-domain structure mapping, and further defines selective mapping of two (or more) input spaces onto an additional space--referred to as \emph{the blend}--which has its own emergent properties realizable by ``running'' or ``elaborating'' the blend \citep{the_way_we_think, concept_blending_comp_fr}. Both analogy and blending are implicated in the creation of new concepts via novel abstractions \citep{Marchiori2024WhatDisruption, the_way_we_think, gentner_smith, mitchell2021abstraction}.

\begin{wrapfigure}{r}{0.5\textwidth}

\centering
\includegraphics[width=0.48\textwidth]{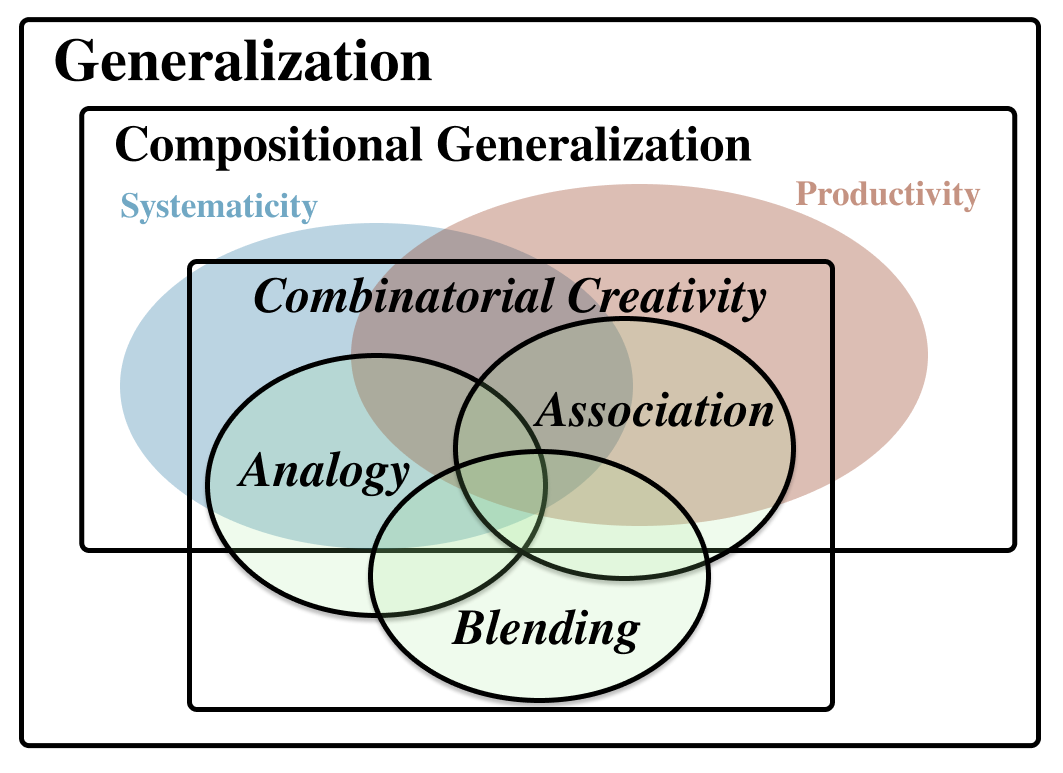}
\caption{\textbf{Conceptual relationships between combinatorial creativity and compositional generalization.} Combinatorial creativity overlaps significantly with compositional generalization, though the two abilities are not identical. Analogy involves structure mapping between domains, which draws upon the systematicity to combine known parts and rules; association involves connecting entities in distant ways, often making longer connections between concepts than those made in training; while blending overlaps the least, involving abstraction ability not typically associated with compositionality.}
\label{fig:generalization_taxonomy}
\end{wrapfigure}

\textbf{Combinatorial Creativity and Compositional Generalization} \quad Among the five types of generalization studied in NLP research \citep{generalization_research_nlp}, \emph{combinatorial creativity} most closely resembles \emph{compositional generalization}.  Broadly, compositionality is a linguistic principle that the meaning of a complex expression is a function of the meaning of its parts and the way they are combined \citep{cogs, systematicity}, and compositional generalization is typically divided into one of five types: systematicity, productivity, substitutivity, localism, and overgeneralization \citep{comp_gen_survey, compositionality_decomposed}. As shown in \Cref{fig:generalization_taxonomy}, combinatorial creativity and compositional generalization are largely overlapping abilities, with association overlapping mostly with productivity (length generalization), analogy overlapping mostly with systematicity (the ability to recombine known parts and rules), and blending overlapping the least, as it involves abstraction and emergent elaboration not typically associated with compositional generalization. More broadly, combinatorial creativity, association, analogy, and blending are not fully subsumed by compositional generalization because creativity involves combining ideas in \emph{unfamiliar and novel ways}, requiring open-ended evaluation uncharacteristic of compositional generalization tasks (cf. \citet{scan} and \citet{cogs}).

\paragraph{Related Work} Although there is no LLM benchmark for combinatorial creativity, previous works have each studied association, analogy, and blending separately. For example, the most closely related associative creativity benchmark over real-world entities is \textsc{Create} \citep{wadhwa2026createtestingllmsassociative}, which assesses the ability to produce diverse and specific connections between entities with known relationships. However, \textsc{Create} does not score output originality---a necessary property of creative products \citep{maher2010, boden, bvsr_update}---and indexes more heavily on closed-ended recognition than open-ended discovery ability due to pre-filtering entities with known connectivity. Psychometric tests of associative creativity, like the Divergent Association Test (DAT) and its variants \citep{Olson2021NamingCreativity, nakajima2026divergentcreativityhumanbasedevaluation, PACE,bellemare2026divergent}, have also been studied extensively in LLMs, but often show poor external validity in predicting creative achievement on real-world tasks like scientific ideation, creative writing, and divergent thinking, casting doubt on their efficacy as psychometric instruments for LLMs \citep{schapiro2026assessingcreativitylargelanguage}. Similarly, \citet{roll_the_dice} studied the effect of pre-training objectives on creativity through synthetic algorithmic tasks loosely inspired by combinatorial creativity, but such tasks do not constitute a benchmark for LLMs. Next, there are numerous closed-ended tasks or benchmarks assessing analogical \emph{reasoning}, the ability for LLMs to recognize or complete known analogies \citep{chen2022kar,lewis2024evaluating,wijesiriwardene2023analogical}, yet comparatively few large-scale studies of analogical \emph{creativity}, the ability for LLMs to discover original analogies and employ them to meet task demands. The most closely related works are small-scale studies examining whether LLMs can generate original analogical mappings to enhance performance on design tasks \citep{liu2026assessingeffectcrossdomainmapping,Kokate_Onkar_2026} or scientific ideation \citep{shen2026unlockingllmcreativityscience}. Lastly, while no large-scale conceptual blending benchmark for LLMs exists, \citet{wang2023popblends} and \citet{song2025peeled} have studied or proposed methods to enhance conceptual blending ability at smaller scale. Most similar to our work, \citet{song2025peeled} study conceptual combination of noun phrases, finding that LLMs (as of 2025) struggle to generate noun phrases with emergent properties (such as the ``white color'' emerging when \texttt{peeled} + \texttt{apple} are combined into ``\texttt{peeled apple}'').
\begin{table}[t]
\centering
\captionof{table}{\textbf{Example outputs on each task}. Colors indicate the domain of each input (\textcolor{domDem}{$\boldsymbol{u}$}, \textcolor{domBank}{$\boldsymbol{v}$}), the \textcolor{newConcept}{\textbf{invented entity}}, and \textcolor{emergentCol}{\textbf{emergent structure}} of the blend. Images are for illustration purposes only.}
\label{tab:examples}
\vspace{2pt}
% Auto-generated by make_showcase_table.py -- do not edit by hand.
% Fragment (no float wrapper, no caption), in the format of tab_examples.tex: \input inside a figure/table float.
\providecommand{\provlogo}[1]{\raisebox{-0.15ex}{\includegraphics[height=0.85em]{media/logos/#1}}\,}
\definecolor{domA}{HTML}{3F6F8E}\definecolor{domB}{HTML}{765030}\definecolor{invCol}{HTML}{22394F}\definecolor{emCol}{HTML}{B74D34}
{\scriptsize
\renewcommand{\arraystretch}{1.3}
\renewcommand\tabularxcolumn[1]{m{#1}}
\begin{tabularx}{\textwidth}{@{}m{1.75cm} X @{\hspace{6pt}}m{0.20\linewidth}@{}}
\toprule
\textbf{Task (model)} & \textbf{Example output} & \textbf{Invention illustration} \\
\midrule
\textbf{(a)} Association \newline (claude-opus-5) & \textbf{Inputs} $(u,v)$: \textcolor{domA}{\emph{Radio}} $\to$ \textcolor{domB}{\emph{The Milky Way}}. \newline
\textbf{Path} $(p)$: (\textcolor{domA}{Radio}, aired hit, Vincent (song)) $\to$ (Vincent (song), tribute to, Vincent van Gogh) $\to$ (Vincent van Gogh, painted, The Starry Night) $\to$ (The Starry Night, swirls depict, \textcolor{domB}{The Milky Way}). & N/A \\
\midrule
\textbf{(b)} Analogy \newline (claude-fable-5) & \textbf{Inputs} $(u,v)$: \textcolor{domA}{\emph{The blue whale}} $::$ \textcolor{domB}{\emph{The mattress}}. \newline
\textcolor{domA}{\textbf{Path a} $(p_u)$: (The blue whale, covered by, epidermis), (epidermis, hosts, barnacles), (barnacles, feed on, plankton)}. \newline
\textcolor{domB}{\textbf{Path b} $(p_v)$: (The mattress, covered by, ticking), (ticking, hosts, dust mites), (dust mites, feed on, skin flakes)}. \newline
\textbf{Source} $(\varphi)$: \textcolor{domB}{vacuum cleaner} (from The mattress, no counterpart in The blue whale). \newline
\textbf{Invention} $(h \gets M[\varphi])$: \textcolor{invCol}{\textbf{whale groomer drone}}. \newline
\textbf{Source to Target Mapping} $(M^{u, v})$: \newline
(\textcolor{domB}{vacuum cleaner}, removes, \textcolor{domB}{dust mites}) $\mapsto$ (\textcolor{invCol}{whale groomer drone}, removes, \textcolor{domA}{barnacles}) \newline
(\textcolor{domB}{vacuum cleaner}, cleans, \textcolor{domB}{ticking}) $\mapsto$ (\textcolor{invCol}{whale groomer drone}, cleans, \textcolor{domA}{epidermis}) \newline
%(\textcolor{domB}{vacuum cleaner}, suctions, \textcolor{domB}{skin flakes}) $\mapsto$ (\textcolor{invCol}{whale groomer drone}, suctions, \textcolor{domA}{plankton}) \newline
(\textcolor{domB}{vacuum cleaner}, maintains, \textcolor{domB}{The mattress}) $\mapsto$ (\textcolor{invCol}{whale groomer drone}, maintains, \textcolor{domA}{The blue whale}) & \includegraphics[width=\linewidth]{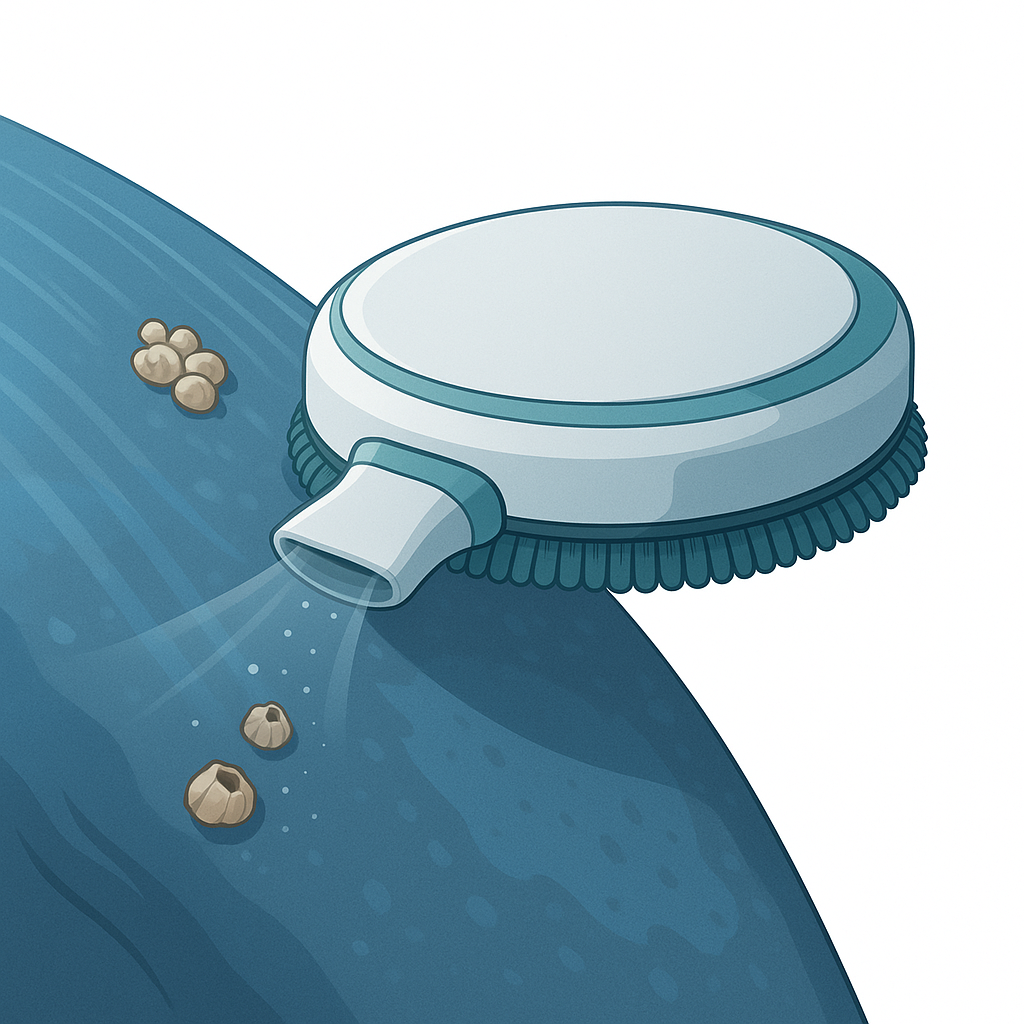} \\
\midrule
\textbf{(c)} Blending \newline (claude-opus-4.5) & \textbf{Inputs} $(u,v)$: \textcolor{domA}{\emph{Adam Smith}} $+$ \textcolor{domB}{\emph{Bacteria}}. \newline
\textbf{Generic Space} $(g)$: autonomous agents whose local signals generate collective order without central control. \newline
\textbf{Invention} $(b)$: \textcolor{invCol}{\textbf{market colony}}. \newline
\textbf{Blended Space} $(c')$, each row tagged by its projection: \newline
(\textcolor{invCol}{market colony}, coordinates through, local signals)~\textcolor{invCol}{[uv]} \newline
(\textcolor{invCol}{market colony}, produces, emergent efficiency)~\textcolor{invCol}{[uv]} \newline
(\textcolor{invCol}{market colony}, specializes via, division of labor)~\textcolor{domA}{[u]} \newline
(\textcolor{invCol}{market colony}, adapts through, replication)~\textcolor{domB}{[v]} \newline
\textcolor{emCol}{(\textcolor{invCol}{market colony}, evolves its, exchange protocols)}~\textcolor{emCol}{[\emph{emergent}]} \newline
\textcolor{emCol}{(\textcolor{invCol}{market colony}, spreads, practices contagiously)}~\textcolor{emCol}{[\emph{emergent}]} & \includegraphics[width=\linewidth]{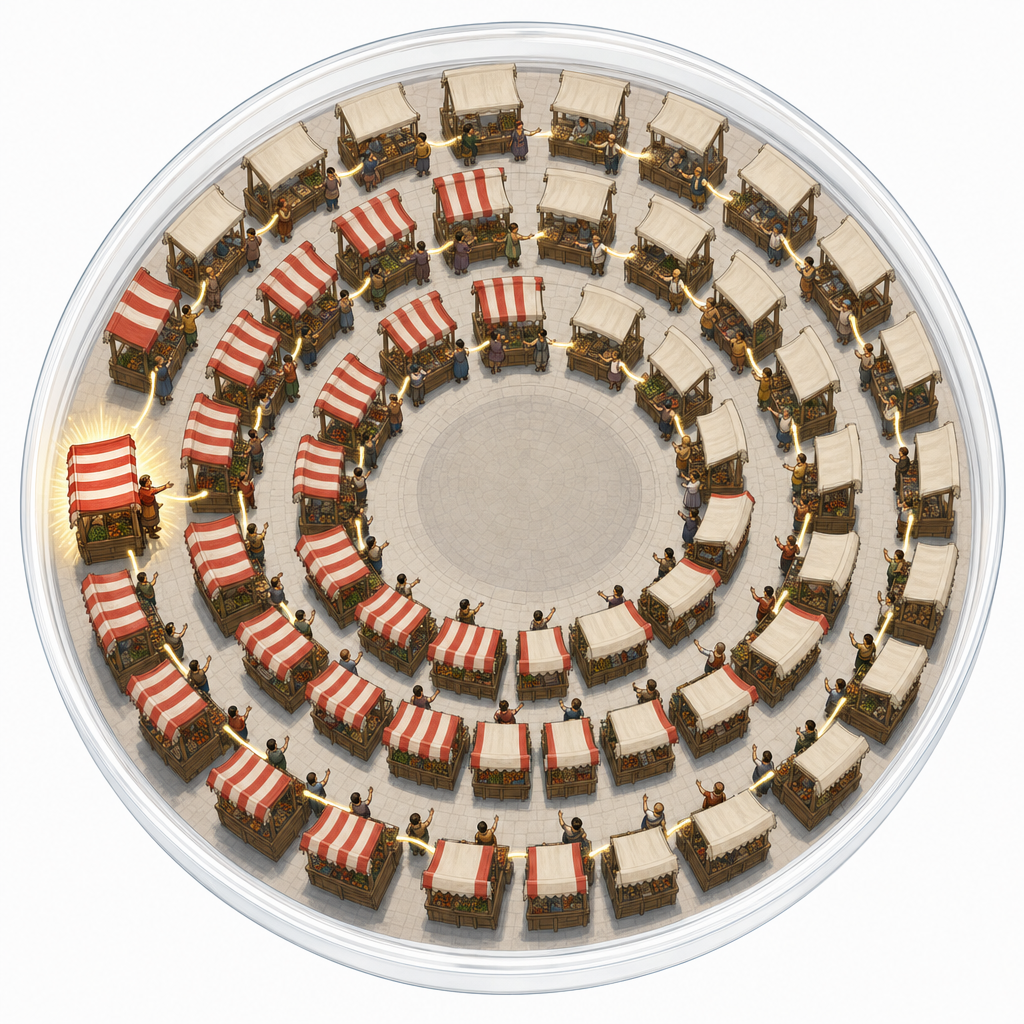} \\
\bottomrule
\end{tabularx}
}

\end{table}

\section{Kombine: An Open-Ended Combinatorial Creativity Benchmark} \label{sec:benchmark}
Here, we introduce a benchmark for combinatorial creativity that assesses its predominant forms---\textbf{association}, \textbf{analogy}, and \textbf{blending}---in an open-ended way. Unlike prior benchmarks which assess closed-ended recognition of known combinations, each task on \textsc{Kombine} involves discovering an association, analogy, or blend between a random pair of input entities $u, v$ with no known relationship (\Cref{tab:examples}). This makes our benchmark more representative of real-world discovery conditions in which novel connections relative to existing human knowledge are the primary desideratum \citep{Swanson1986UndiscoveredKnowledge}.\footnote{Though see the distinction between historical vs. personal creativity acknowledged in \Cref{sec:conclusion}.} Using these discovered relationships, models are then tasked to \emph{invent novel concepts} by re-combining the parts of existing ones, employing the structure mappings afforded by analogies and blends. We start by providing shared notation and setup, then formally define each task and its scoring, and lastly detail our experimental setup.

\paragraph{Preliminaries} Combinatorial creativity occurs over a space of entities $\mathcal{C}$ and relations $\mathcal{R}$. An entity-relation-entity \emph{triple} is a tuple $t = (c, r, c') \in \mathcal{C} \times \mathcal{R} \times \mathcal{C}$ whose truth or falsity is determined by a judge $F : \mathcal{C} \times \mathcal{R} \times \mathcal{C} \to \{0,1\}$. Consistent with a large body of literature in psychometrics and computational creativity that assesses semantic divergence via the embedding distance between elements \citep{Olson2021NamingCreativity, Johnson2023DivergentModeling, explaining_creative_artifacts}, we define a text embedding function $\phi$ and let $d^\phi_\text{cos} : \mathcal{C} \times \mathcal{C} \to [0, 1]$ denote its induced cosine distance measure. For an added element $e$ of a response to inputs $u, v \in \mathcal{C}$ (i.e., any $e \in \mathcal{R} \cup \mathcal{C} \setminus \{u, v\}$), $\rho(e)$ is its mean cosine distance to its $m$ nearest neighbors among the distinct added elements that all models (or humans, as in \Cref{subsec:human_results}) produced for the same item, so that an element is original when nothing semantically close to it was produced by any model (or human)---including that same participant's previous items for association, so originality implicitly captures diversity \citep{hivemind, zhang2025noveltybenchevaluatinglanguagemodels}.\footnote{By default, we use \texttt{all-MiniLM-L6-v2} \citep{reimers2019sentencebert} for $\phi$, but \Cref{app:embedding_robustness} reports robustness across varying embedding models. Moreover, originality uses $m = 5$ nearest neighbors.}  An artifact's originality is then the mean of $\rho$ over all its added elements. Each task is scored according to utility, surprise, and originality (novelty), which are established criteria for creative products in the psychology literature \citep{maher2010, creativity_in_science, boden}. On all three tasks, utility is a binary indicator of whether the task-specific constraints are met, consistent with psychometric practice that scores this dimension as either satisfied or not \citep{maher2010, nakajima2026divergentcreativityhumanbasedevaluation}. On analogy and blending, we also assess the \textbf{\emph{emergent creativity}} of invented concepts according to their originality $O$, coherence and feasibility $V$, and semantic integration quality $Q$. Both $V$ and $Q$ are assessed using LLM judges whose prompts are given in \Cref{app:judge_prompts}. We now formally define each task and its scoring.

\subsection{Association Task}
The association task assesses the ability to connect two distant entities otherwise believed to be unrelated, by discovering a surprising, original, and factual path of triples between them. On association, models are tasked to provide as many distinct paths as possible, and each dimension is averaged across all paths.

\begin{taskbox}[def:association]{Association}
Given inputs $u, v$, an \emph{association} is a path of triples $p=\big((u,r_1,c_1),\dots,(c_{n-1},r_n,v)\big)$ connecting them ($c_0=u$, $c_n=v$). It is \emph{useful} $U_\text{as}(p) := \prod_{t\in p} F(t) \in \{0, 1\}$ only when every triple is factual, its \emph{surprise} $S_\text{as}(p) := \tfrac{1}{n-1}\sum_{i} d^\phi_{\cos}(c_{i-1},c_i)$ is the average semantic distance between consecutive entities in the path, and its \emph{originality} $O_\text{as}(p) := \tfrac{1}{|E_p|}\sum_{e\in E_p}\rho(e)$ is the mean distance of its added elements $E_p$ from the response pool.
\end{taskbox}
\vspace{2pt}

\subsection{Analogy Task}
The analogy task involves discovering relational structure shared by the two unrelated inputs, and then using this mapping to invent a new concept relevant to one of the inputs. Formally, a pair of length-$n$ paths $p_u=\big((u,r_1,a_1),\dots,(a_{n-1},r_n,a_n)\big)$ and $p_v=\big((v,q_1,b_1),\dots,(b_{n-1},q_n,b_n)\big)$ ($a_0=u$, $b_0=v$) that share a relation sequence ($r_i\equiv q_i$) define a \emph{structure mapping} $M^{u, v} : \mathcal{C} \to \mathcal{C}$ from each entity of $p_u$ to its counterpart in $p_v$, $M(a_i):=b_i$, which extends to triples and to sets or paths of triples $T$ according to $M[(c,r,c')]:=(M(c),\,r,\,M(c'))$ and $M[T]:=\{M[t]:t\in T\}$. Consistent with structure mapping theory, the mapping is semantically valid only if every triple in each path is factual and the two paths satisfy relational similarity.

\begin{taskbox}[def:analogy]{Analogy}
Given inputs $u, v$, an \emph{analogy} $a = (p_u \cup p_v, h)$ consists of a pair of paths $p_u, p_v$ that define a structure mapping $M$, and a concept $h:=M[\Phi]$ invented by projecting a source domain entity $\varphi$ and its structure $\Phi := \{(\varphi, r_j, s_j)\}_{j=1}^k$ through the mapping. Its \emph{utility} $U_\text{an}(a) := \prod_{i=1}^n F(p_u[i]) F(p_v[i]) \cdot \prod_{i=1}^{n} \mathbb{I}[r_i \equiv q_i] \in \{0, 1\}$ holds when the mapping $M$ is semantically valid, its \emph{surprise} $S_\text{an}(a) := \tfrac{1}{n+1}\sum_{i} d^\phi_{\cos}(a_i,b_i)$ is the average distance between entities in corresponding roles of each path, and its \emph{originality} $O_\text{an}(a)$ is defined as for association over its added elements $E_a$. Lastly, the \textbf{\emph{emergent creativity}} of $h$ is scored by the invented concept's originality $O(h)$, utility $V(h) \in \{0, 1\}$, and semantic integration quality $Q(h) \in \{0, 1\}$, which indicates whether the mapping was actually used to construct $h$.
\end{taskbox}
\vspace{2pt}

\subsection{Blending Task}
Lastly, the blending task involves discovering abstract structure both inputs instantiate (referred to by \citet{the_way_we_think} as a \emph{generic space}) and fusing them into a new space which inherits structure from both inputs. Blends allow one to ``elaborate'' in this blended space to describe emergent properties, such as a market colony that ``evolves its exchange protocols'' (\Cref{tab:examples}). Following \citet{the_way_we_think}, a blend is \emph{single-scope} when both inputs contribute projected structure, \emph{double-scope} when both intersect on at least one shared property, and \emph{double-scope emergent} when a double-scope blend also elicits an emergent property.

\begin{taskbox}[def:blend]{Blending}
Given inputs $u, v$, a \emph{blend} $a = (g, c')$ consists of a generic space $g$ that both inputs instantiate and a blended space describing an invented concept $c' := M_u[\Phi_u] \cup M_v[\Phi_v] \cup \Delta$, into which mappings $M^{u, c'}, M^{v, c'}$ selectively project triples $\Phi_u := \{(u, r_j, s_j)\}_j$ and $\Phi_v := \{(v, q_j, t_j)\}_j$ true of each input, and where $\Delta := \{t \in c' : u, v \not\models t\}$\footnotemark{}\xdef\fnEntail{\arabic{footnote}} is emergent structure true of the blend but of neither input. Its \emph{utility} $U_\text{bl}(g) := G(g) \in \{0,1\}$ is scored by a judge $G$ of whether the generic space $g$ fits both inputs, its \emph{surprise} $S_\text{bl}(g) := \tfrac{1}{2}\big(d^\phi_{\cos}(u,g) + d^\phi_{\cos}(v,g)\big)$ is the mean distance of $g$ from the two inputs, and its \emph{originality} $O_\text{bl}(g) := \rho(g)$ is the distance of $g$ from the generic spaces in other responses to the same item.\footnotemark{}\xdef\fnGeneric{\arabic{footnote}} The \textbf{\emph{emergent creativity}} of $c'$ is scored by its originality $O(c')$, utility $V(c') \in \{0, 1\}$ (whether each triple $t \in c'$ is feasible or coherent), and integration quality $Q(c') := 1$ if $M_u[\Phi_u], M_v[\Phi_v] \neq \emptyset$ (single-scope), $2$ if additionally $M_u[\Phi_u] \cap M_v[\Phi_v] \neq \emptyset$ (double-scope), and $3$ if additionally $\Delta \neq \emptyset$ (double-scope emergent).
\end{taskbox}

\footnotetext[\fnEntail]{Here, $\models$ refers to the textual entailment operator (i.e., ``text T entails hypothesis H if a human reading T would infer H is most likely true.'') \citep{dagan2005pascal}. In this case, we refer to opposite entailment (e.g., a human or LLM judge reading T would \emph{not} infer H is true).}
\footnotetext[\fnGeneric]{Note that the generic space is a text piece, so originality here is computed over the sentence embedding, rather than over relations and concepts individually as in analogy and association.} 

\subsection{Inventive Multiples} Given that LLMs have a documented tendency to homogenize on open-ended tasks \citep{hivemind}, we also study the degree of conceptual overlap between inventions made from the analogy and blending tasks. By analogy to \emph{scientific multiples} in the science of science \citep{simonton_discovery_invention_as_combinatorial, creativity_in_science}, we refer to near-identical inventions (made by different models) given the same inputs $(u, v) \in \mathcal{C}^2$ as \emph{inventive multiples}, and formally define them as follows.

\begin{definition}[$\tau$-Inventive Multiples] \label{def:inventive_multiples}
    Let $a, a' \in \mathcal{C}$ be two concepts invented via analogy or blending given the same input entities $u,v \in \mathcal{C}$. Assume $a, a'$ are described by property sets $P = \{p_j \}_{j=1}^k$ and $P' = \{p'_j\}_{j=1}^{k'}$, respectively, where each property $p_j, p_j' \in \mathcal{R} \times \mathcal{C}$ is a relation-object tuple (e.g., (\texttt{can be, traded})). Given a similarity threshold $\theta > 0$, two properties are semantically indistinguishable, written $p \approx p'$, whenever $\cos\big(\phi(p), \phi(p')\big) \ge \theta$. Then, $a$ and $a'$ are $\tau$\textbf{-inventive multiples} if at least $\tau \in \mathbb{N}$ distinct properties of $a$ (i.e., ($p_1, \dots, p_\tau$)) have unique, semantically indistinguishable counterparts in $a'$ (i.e., $(p'_1, \dots, p'_\tau)$ such that $p_i \approx p_i', \forall i \le \tau$). 
\end{definition}

\subsection{Experimental Setup for LLMs}
We study a set of 35 LLMs, comprising a mix of the latest frontier systems (e.g., GPT-6 Astra, Claude Fable 5.1) and smaller, open-weight models. All models are sampled at temperature $T = 0.9$ with top-p and top-k parameters disabled and reasoning level left at default. To instantiate the space of entities $\mathcal{C}$, we randomly sample entities from the Wikidata knowledge graph, restricted to recognizable entities with at least $k=78$ sitelinks, as explained in \Cref{app:entity_pool}. The full list of entities is given in \Cref{tab:entity_pool}. We use \texttt{Claude Haiku 4.5} as the LLM factuality judge $F$, and a panel of three LLMs (\texttt{Claude Haiku 4.5}, \texttt{gpt-o3}, and \texttt{GPT-5.4}) for all other analogy and blending judges, with prompts given in \Cref{app:judge_prompts}. Analogy and blend judges achieve fair-to-good reliability, and they agree with human judgment 75\% of the time on items for which the three-judge LLM panel is unanimous. Furthermore, restricting the leaderboard to items where the LLM judge panel is unanimous leaves the model ranking almost entirely the same (see \Cref{app:reliability}).

\subsection{Human Data Collection}
To study human performance on each of the three tasks, we recruit $120$ online English-speaking adults from Prolific, retaining a final sample of $N = 75$ participants after screening participants for AI usage ($32$ male, $40$ female, $1$ non-binary, $1$ genderfluid, and $1$ preferred not to say). The mean age among participants was $40.1$ years (SD = $12.7$, range=$[20,78]$). Each participant completed five association items, five analogy items, and five blending items, with entity pairs drawn from the same pool as in LLM experiments (\Cref{tab:entity_pool}). Each task was introduced by a worked example and a short comprehension check, and opened by an easy warm-up item (\texttt{Life :: A journey} for analogy and \texttt{Breakfast + Lunch} for blending). The human study platform is shown in \Cref{fig:platform_association,fig:platform_analogy,fig:platform_blending}. Human responses were scored by the same judges and scoring dimensions as LLM responses, and any item a participant skipped receives a score of zero on every dimension.

\begin{table}[t]
\centering
% Auto-generated by make_composite_table.py -- do not edit by hand.
% Fragment (no float wrapper): \input inside a table float so it stays directly above the
% examples table (\captionof needs the caption package).
\captionof{table}{\textbf{Top systems by combinatorial creativity.} The highest-scoring systems on \textit{Overall} and on each task, as the equal-weight composite of each scoring dimension in \Cref{sec:benchmark}.}
\label{tab:leaderboard}
\providecommand{\provlogo}[1]{\raisebox{-0.15ex}{\includegraphics[height=0.85em]{media/logos/#1}}\,}
{\tiny
\setlength{\tabcolsep}{5pt}
\resizebox{\textwidth}{!}{%
\begin{tabular}{rllll}
\toprule
Rank & \textbf{Overall} & \textbf{Association} & \textbf{Analogy} & \textbf{Blending} \\
\midrule
1 & \provlogo{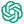}gpt-5.6-sol~(55.2) & \provlogo{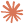}claude-opus-4.6~(51.4) & \provlogo{claude}claude-opus-4.6~(63.1) & \provlogo{openai}gpt-5.6-sol~(64.8) \\
2 & \provlogo{openai}gpt-6-astra~(54.4) & \provlogo{openai}gpt-5.2~(50.3) & \provlogo{openai}gpt-6-astra~(61.1) & \provlogo{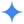}gemini-3.1-pro~(59.5) \\
3 & \provlogo{claude}claude-opus-4.6~(52.0) & \provlogo{googlegemini}gemini-2.5-flash~(48.3) & \provlogo{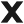}grok-4.6~(60.1) & \provlogo{openai}gpt-6-astra~(56.5) \\
4 & \provlogo{googlegemini}gemini-3.1-pro~(48.7) & \provlogo{claude}claude-opus-4.5~(48.0) & \provlogo{openai}gpt-5~(58.6) & \provlogo{googlegemini}gemini-3.7-flash~(52.9) \\
5 & \provlogo{claude}claude-fable-5~(48.1) & \provlogo{openai}gpt-5-mini~(47.9) & \provlogo{xai}grok-4.5~(55.4) & \provlogo{claude}claude-fable-5~(45.7) \\
6 & \provlogo{xai}grok-4.6~(47.6) & \provlogo{xai}grok-4.6~(47.8) & \provlogo{openai}gpt-5.6-sol~(55.4) & \provlogo{claude}claude-fable-5.1~(43.3) \\
7 & \provlogo{openai}gpt-5~(47.4) & \provlogo{xai}grok-4.5~(47.5) & \provlogo{claude}claude-sonnet-4.6~(54.3) & \provlogo{claude}claude-opus-4.6~(41.4) \\
8 & \provlogo{xai}grok-4.5~(46.9) & \provlogo{claude}claude-fable-5~(47.3) & \provlogo{claude}claude-fable-5~(51.3) & \provlogo{openai}gpt-5.2~(39.1) \\
% 9 & \provlogo{googlegemini}gemini-3.7-flash~(46.4) & \provlogo{openai}gpt-5~(46.2) & \provlogo{openai}gpt-5.2~(49.0) & \provlogo{claude}claude-sonnet-4.5~(38.3) \\
% 10 & \provlogo{openai}gpt-5.2~(46.1) & \provlogo{openai}gpt-6-astra~(45.6) & \provlogo{claude}claude-opus-4.5~(47.4) & \provlogo{xai}grok-4.5~(37.6) \\
\midrule
35 & \provlogo{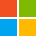}phi-4~(17.4) & \provlogo{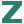}glm-4.5-air~(19.6) & \textbf{\textit{Humans}}~(19.1) & \textbf{\textit{Humans}}~(7.4) \\
36 & \textbf{\textit{Humans}}~(13.7) & \textbf{\textit{Humans}}~(14.7) & \provlogo{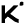}kimi-k2~(17.0) & \provlogo{microsoft}phi-4~(4.4) \\
\bottomrule
\end{tabular}}
}

\vspace{0.2em}

% Generic-space failures grid; shares the benchmark's float with the leaderboard (tab:leaderboard).
\includegraphics[width=0.82\linewidth]{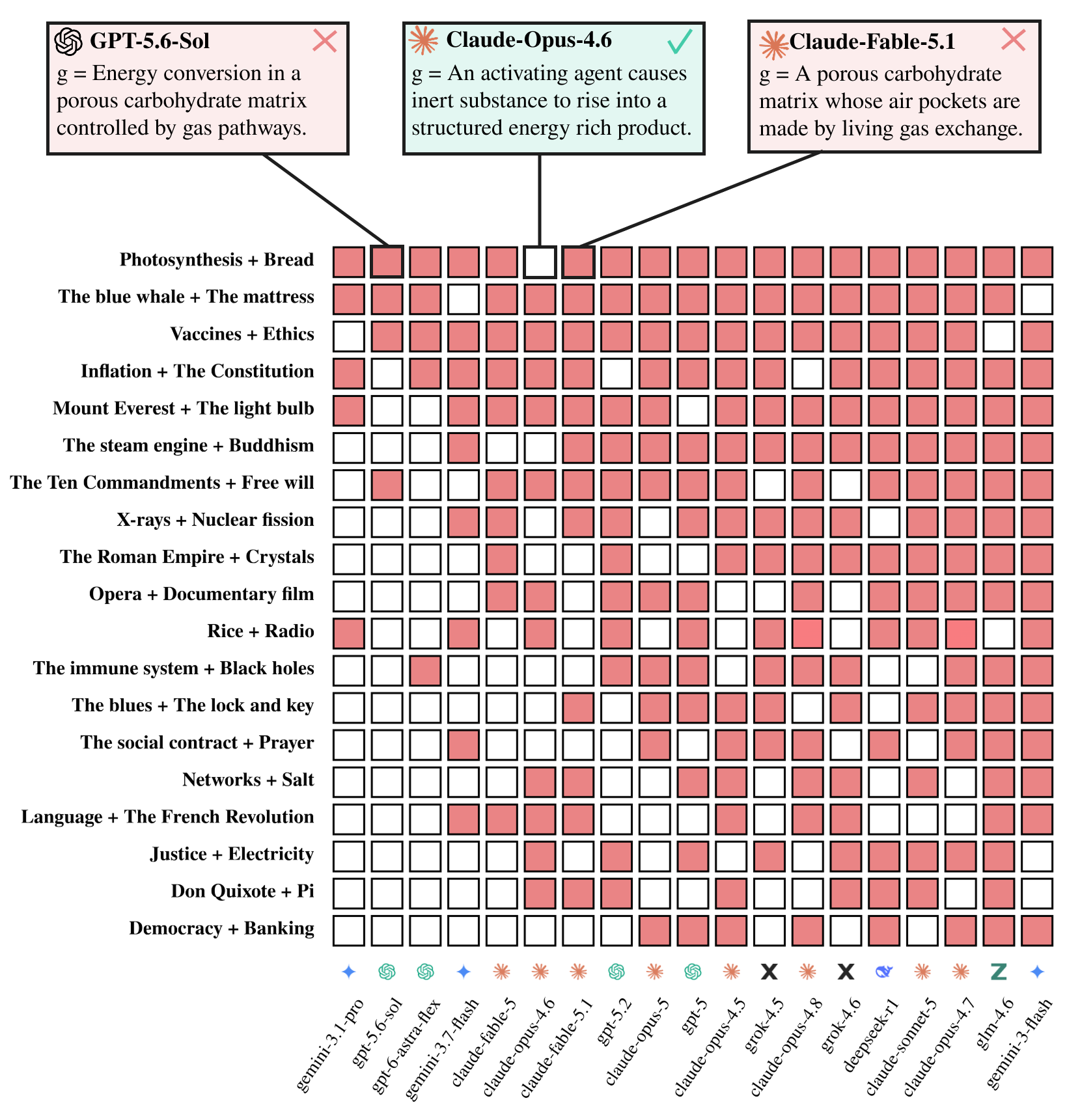}
\captionof{figure}{\textbf{Blending generic space failures among frontier models.} A red cell marks an item on which a model's generic space was invalid, often due to applying only to one input. For example, \texttt{GPT-5.6-Sol's} generic space is subtly wrong, as yeast does convert energy inside a porous starch matrix, but in bread the gas spaces are a product of that conversion, not something that \emph{controls} it. Similarly, \texttt{Fable-5.1}'s generic space is invalid because it does not apply to plants going through photosynthesis, whose air spaces are not made by gas exchange. 
Although blending is difficult for LLMs, no row is fully red, i.e., every entity pair had \emph{at least one model} obtain a valid generic space.}
\label{fig:abstraction_failure}

\end{table}

\section{Results and Key Findings} \label{sec:results}

\subsection{Creative Strengths and Weaknesses of LLMs} \label{subsec:llm_results}

% Facet correlations: back in the results section as a right wrapfigure for the preprint (2026-09-26).
\begin{wrapfigure}{r}{0.4\linewidth}
\centering
\includegraphics[width=\linewidth]{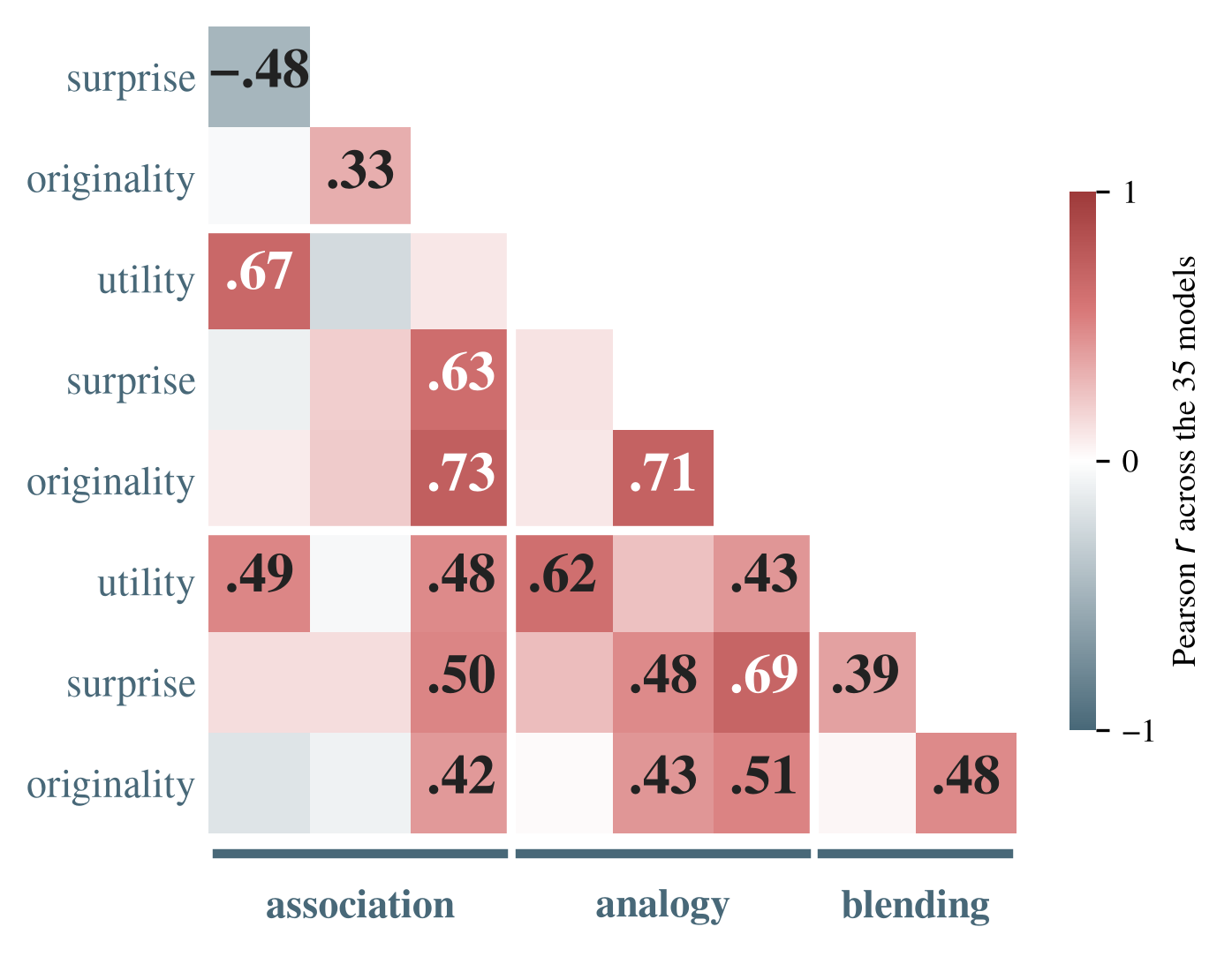}
\captionof{figure}{\textbf{Facet correlations across the 35 models,} restricted to the base dimensions with invention scoring excluded. Only cells with $p < 0.05$ are labeled.}
\label{fig:facet_corr}
\end{wrapfigure}

\Cref{tab:leaderboard} reports the top eight models overall, and on each of the three tasks (see \Cref{tab:leaderboard_full} for the full ranking of all 35 models). Each task score is the mean of that task's scoring dimensions, reported as a fraction of each dimension's maximum, and the overall score is the mean of the three task scores. Importantly, on artifacts that fail utility, every dimension is set to zero. Although association and analogy correlate at $r = 0.71$ ($p < 0.0001$), analogy and blending at $r = 0.70$ ($p < 0.0001$) and blending and association at $r = 0.54$ ($p < 0.001$), the top performing models are task-specific, with \texttt{claude-opus-4.6} leading on association and analogy, but \texttt{gpt-5.6-sol} leading on blending (\Cref{tab:leaderboard}). The rankings are also robust to the choice of embedding model, as detailed in \Cref{app:embedding_robustness}.

\textbf{Association and analogy are easiest for LLMs, while blending is the hardest} \quad Averaged over the 35 LLMs, models produce a factual association 59.0\% of the time (\Cref{tab:full_association}), a valid analogy ($U_\text{an}(a) = 1$ in the \hyperref[def:analogy]{Analogy Task}) 53.4\% of the time (\Cref{tab:full_analogy}), and a valid blend ($U_\text{bl}(a) = 1$ in the \hyperref[def:blend]{Blending Task}) 43.4\% of the time (\Cref{tab:full_blending}).\footnote{Note, however, that this aggregate statistic is not true of every model, as \texttt{gpt-5.6-sol} scores 64.8\% on blending but only 55.4\% on analogy.} 

\textbf{LLM abstraction ability is the key bottleneck in blending} \quad 
Although blending is most difficult for LLMs on average, all of the entity pairs $(u, v)$ had at least one model discover a valid blend (see \Cref{fig:abstraction_failure}), so no item was impossible to blend. For \texttt{(Photosynthesis, Bread)}, Opus 4.6 gave the only valid generic space ($g = $~\texttt{An activating agent causes inert substance to rise into a structured energy rich product}), while all other frontier models failed (e.g., Fable 5.1 said $g =$~\texttt{a porous carbohydrate matrix whose air pockets are made by living gas exchange}, which only applies to bread). Overall, as shown in \Cref{tab:full_blending}, models can find a generic space only $43.4\%$ of the time on average, but conditioned on finding a valid generic space, models can describe a coherent blended concept $97.1\%$ of the time, a finding we refer to as the \emph{abstraction bottleneck}. One plausible explanation for why this occurs is lack of experiential knowledge, as human neuro-imaging studies implicate mental imagery and episodic simulation in creative association and conceptual combination \citep{benedek2020elements}.

\begin{table}[t]
\centering
\captionof{table}{\textbf{Examples of two inventive multiples.} Semantically indistinguishable properties are placed side-by-side and highlighted. \emph{Operators}: Blending (\textbf{+}), Analogy (\textbf{::}).}
\label{tab:multiples_examples}
% Auto-generated by make_multiples_examples_table.py -- do not edit by hand.
% Fragment (no float wrapper, no caption): \input inside a table float; the caption and label
% live in the section file next to the \input.
\providecommand{\provlogo}[1]{\raisebox{-0.15ex}{\includegraphics[height=0.85em]{media/logos/#1}}\,}
\definecolor{multRow}{HTML}{486878}
\definecolor{rowgrey}{HTML}{F2F2F2}
\scriptsize
\setlength{\tabcolsep}{2pt}
\renewcommand{\arraystretch}{1.05}
% Both panels are typeset once into boxes to measure them, then the shorter one is padded on its last row so the
% two bottom rules end at the same height (Alexi, 2026-09-22: "can we make these figures the same height?").
\newlength{\multpadA}\newlength{\multpadB}\setlength{\multpadA}{0pt}\setlength{\multpadB}{0pt}
\newsavebox{\multboxA}\newsavebox{\multboxB}
\newcommand{\multpanelA}{%
\textbf{\footnotesize \emph{Opera} + \emph{Documentary film}}\\[3pt]
\begin{tabular}{@{}>{\raggedright\arraybackslash}p{0.43\linewidth}@{\hspace{4pt}}c@{\hspace{4pt}}>{\raggedright\arraybackslash}p{0.43\linewidth}@{}}
\toprule
\provlogo{claude}claude-fable-5.1 & & \provlogo{googlegemini}gemini-3.7-flash \\
\textbf{``testimony opera''} & {\scriptsize cos} & \textbf{``docu-opera''} \\
\midrule
\rowcolor{multRow!38} $p_{2}$: intercuts archival footage & $0.87$ & $p'_{2}$: incorporates archival footage \\
\rowcolor{multRow!31} $p_{4}$: derives melody from recorded speech & $0.81$ & $p'_{4}$: derives melody from spoken pitch contours \\
\rowcolor{multRow!26} $p_{3}$: halts action for sung testimony & $0.77$ & $p'_{3}$: delivers sung verbatim testimony \\
\midrule
\textcolor{gray}{$p_{1}$: accompanied by orchestra} & & \textcolor{gray}{$p'_{1}$: utilizes orchestral overtures} \\[\multpadA]
% \textcolor{gray}{$p_{5}$: chorus sings source citations} & & \textcolor{gray}{$p'_{5}$: reveals factual deception via harmonic dissonance} \\
\bottomrule
\end{tabular}}
\newcommand{\multpanelB}{%
\textbf{\footnotesize \emph{Democracy} :: \emph{Banking}}\\[3pt]
\begin{tabular}{@{}>{\raggedright\arraybackslash}p{0.43\linewidth}@{\hspace{4pt}}c@{\hspace{4pt}}>{\raggedright\arraybackslash}p{0.43\linewidth}@{}}
\toprule
\provlogo{claude}claude-fable-5 & & \provlogo{claude}claude-fable-5.1 \\
\textbf{``vote insurance''} & {\scriptsize cos} & \textbf{``legitimacy stress test''} \\
\midrule
\rowcolor{multRow!34} $p_{1}$: protects ballot & $0.84$ & $p'_{3}$: protects votes \\
\rowcolor{multRow!18} $p_{5}$: mandated by election commission & $0.70$ & $p'_{4}$: is overseen by electoral commission \\
\midrule
\textcolor{gray}{$p_{2}$: restores popular will} & & \textcolor{gray}{$p'_{1}$: simulates recall election} \\
\textcolor{gray}{$p_{3}$: triggered by election annulment} & & \textcolor{gray}{$p'_{2}$: measures mandate} \\
\textcolor{gray}{$p_{4}$: prevents voter boycott} & &  \\[\multpadB]
\bottomrule
\end{tabular}}
\savebox{\multboxA}{\begin{minipage}[t]{0.49\linewidth}\centering\multpanelA\end{minipage}}
\savebox{\multboxB}{\begin{minipage}[t]{0.49\linewidth}\centering\multpanelB\end{minipage}}
\ifdim\dimexpr\ht\multboxA+\dp\multboxA\relax>\dimexpr\ht\multboxB+\dp\multboxB\relax
  \setlength{\multpadB}{\dimexpr\ht\multboxA+\dp\multboxA-\ht\multboxB-\dp\multboxB\relax}
\else
  \setlength{\multpadA}{\dimexpr\ht\multboxB+\dp\multboxB-\ht\multboxA-\dp\multboxA\relax}
\fi
\begin{minipage}[t]{0.49\linewidth}\centering\multpanelA\end{minipage}\hfill
\begin{minipage}[t]{0.49\linewidth}\centering\multpanelB\end{minipage}

% The rate of tau-inventive multiples (plot_tau_multiples.py in the analysis repo, from tau_curve in
% inventive_multiples.json): the appendix tau table as a figure. Float-less: shares the benchmark
% section's float with the multiples examples table. Caption drafted in \ai{} for the author.
\par\vspace{14pt}
% Three panels as subfigures so the (a)/(b)/(c) labels are LaTeX subcaptions rather than in-plot titles
% (Alexi, 2026-09-22); the panel files carry no titles and keep the widths of the combined figure.
\begin{subfigure}[t]{0.3175\linewidth}\centering\includegraphics[width=\linewidth]{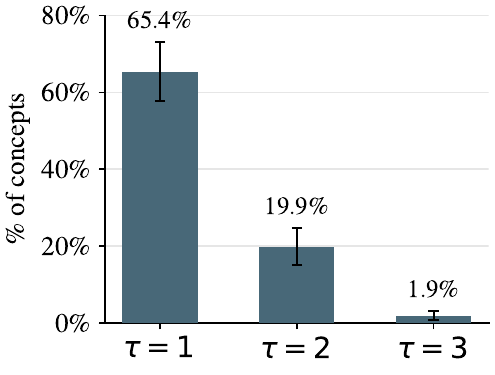}\caption{Concepts in a multiple}\end{subfigure}\hfill
\begin{subfigure}[t]{0.3465\linewidth}\centering\includegraphics[width=\linewidth]{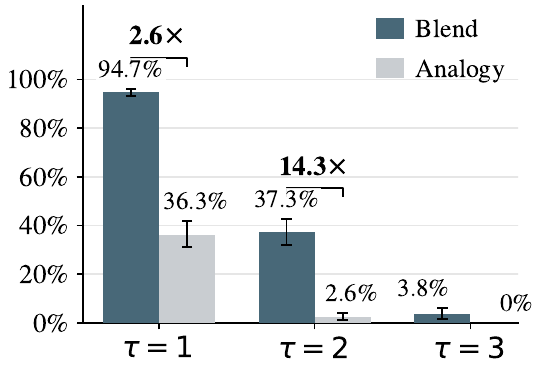}\caption{By task}\end{subfigure}\hfill
\begin{subfigure}[t]{0.3175\linewidth}\centering\includegraphics[width=\linewidth]{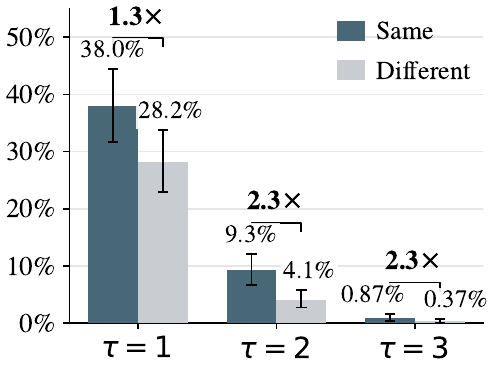}\caption{By family}\end{subfigure}
% Row 2: the per-provider figure (plot_family_multiples.py), merged as (d); back from the appendix for the preprint (2026-09-26).
\par\vspace{4pt}
\begin{center}\begin{subfigure}[t]{0.52\linewidth}\centering\includegraphics[width=\linewidth]{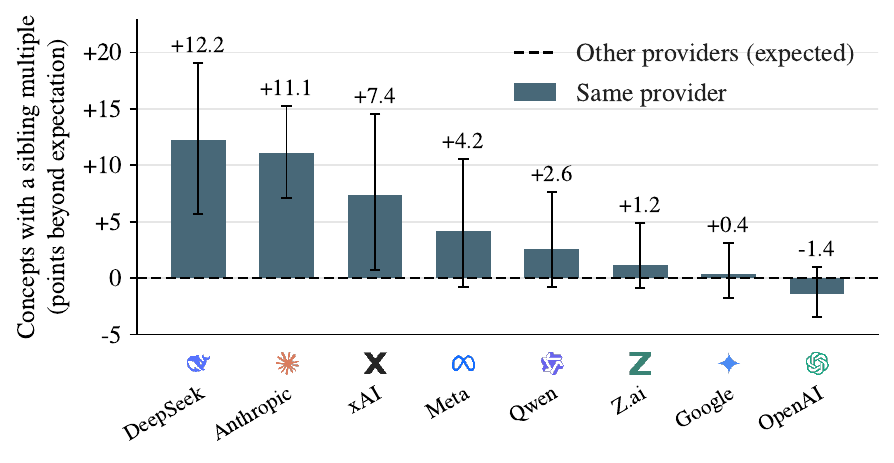}\caption{By provider}\label[subfigure]{fig:family_multiples}\end{subfigure}\end{center}
\captionof{figure}{\textbf{Rate of inventive multiples.} \textbf{(a)} The rate of all 2,070 inventions that belong to at least one $\tau$-multiple, aggregated by task \textbf{(b)} and by model family \textbf{(c)}, with task and family ratios above each $\tau$. Error bars are 95\% confidence intervals from a bootstrap over items. \textbf{(d)} The rate of invention pairs from two models of the same provider that are $\tau = 2$ multiples for every provider with at least two models in the pool. The dashed line is the rate for pairs of models from different providers.}
\label{fig:tau_multiples}

\end{table}

\textbf{Association originality is the best predictor of every other facet} \quad \Cref{fig:facet_corr} showcases facet correlations across all three core scoring dimensions in \Cref{sec:benchmark}. LLMs which find more original ways to connect distant entities otherwise believed to be unrelated, also tend to find more original ($r = 0.73$, $p < 1\mathrm{e}{-4}$) and surprising analogies ($r = 0.63$, $p = 6\mathrm{e}{-5}$), more original ($r = 0.42$, $p = 1\mathrm{e}{-2}$) and surprising blends ($r = 0.50$, $p = 2\mathrm{e}{-3}$), and valid blends more often ($r = 0.48$, $p = 3\mathrm{e}{-3}$). Overall, association originality has the highest mean cross-task correlation of any other facet ($\bar{r} = 0.47$), consistent with Mednick's \citeyearpar{associative_basis} associative theory of creativity---stipulating that the ability to connect distant entities in original ways is central to most types of creativity. As detailed in \Cref{app:embedding_robustness}, these relationships hold across varying embedding model choices. 

\subsection{The Inventive Homogeneity of LLMs} \label{subsec:inventive_homogeneity_llms}
Across all 35 LLMs, a total of 2,070 concepts was invented through analogies and blends, and these concepts were described by 7,868 total (relation, object) properties. To determine which of these are multiples, we set $\theta = 0.674$ as the similarity threshold in \Cref{def:inventive_multiples}, which corresponds to the 99th percentile of cosine similarity over all property pairs. 

\textbf{One in five invented concepts is a multiple} \quad \Cref{fig:tau_multiples}(a) showcases the proportion of invented concepts containing $\tau \in \{1, 2, 3\}$ semantically indistinguishable properties from another model's concept invented for the same item. At $\tau = 2$, 19.9\% of the invented concepts (412/2,070) is a multiple.

\textbf{Models are $3{-}14 \times$ more likely to produce multiples on blends than analogies} \quad As shown in \Cref{fig:tau_multiples}(b), a concept invented by blending is $2.6\times$ more likely to be a $(\tau{=}1)$-inventive multiple than one invented by analogy (94.7\% vs. 36.3\%), and $14.3\times$ more likely to be a $(\tau{=}2)$-inventive multiple (37.3\% vs. 2.6\%).

\paragraph{Models from the same family invent alike} As shown in \Cref{fig:tau_multiples}c, models from the same family are $2.3\times$ more likely to invent ($\tau{=}2$)-multiples than models from different providers (9.3\% vs. 4.1\%). In \Cref{fig:family_multiples}, we report the intra-family multiple rate as a function of model provider. To normalize the rate of multiples among providers with unequal numbers of models in the pool, we compare the rate of inventive multiples within a single provider's $k$ models, relative to that provider's average rate of multiples with $k$ different-provider models in the pool, and plot the difference as a function of provider in \Cref{fig:family_multiples}. DeepSeek and Anthropic models exhibit the greatest intra-family multiple rate at 12.2 and 11.1 percentage points (pp) above baseline, while OpenAI models exhibit the least, converging with each other 1.4pp \emph{less often} than they do with models from other providers.

% Typical human vs. LLM concepts and relations (plot_human_llm_language.py in the analysis repo): whole phrases pooled
% over the three tasks, sized by the log-odds z with an informative Dirichlet prior (Monroe et al. 2008); kept humans vs.
% the 35 LLMs on the 86 shared items. Caption left to the author.
\begin{figure}[t]
    \centering
    \begin{subfigure}[t]{0.495\linewidth}\centering\includegraphics[width=\linewidth]{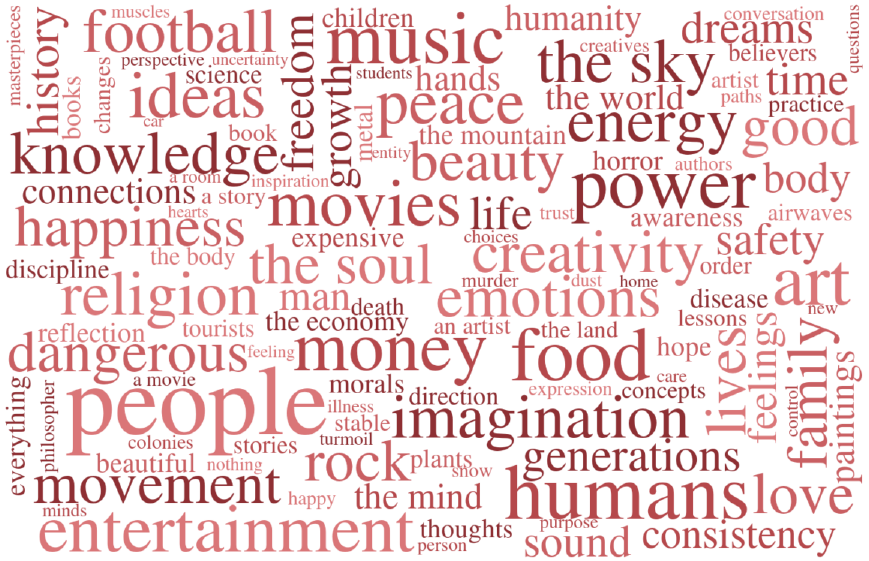}\caption{Human concepts}\end{subfigure}\hfill
    \begin{subfigure}[t]{0.495\linewidth}\centering\includegraphics[width=\linewidth]{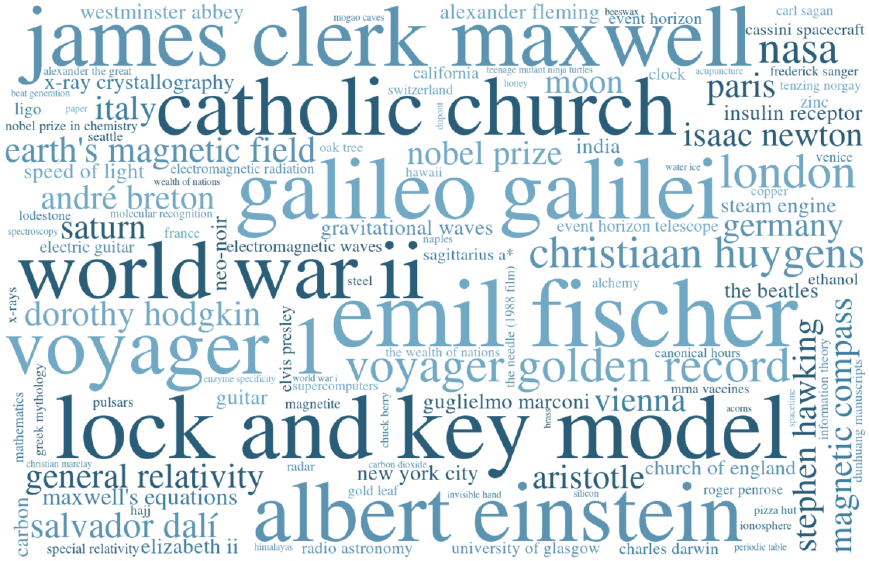}\caption{LLM concepts}\end{subfigure}\\[4pt]
    \begin{subfigure}[t]{0.495\linewidth}\centering\includegraphics[width=\linewidth]{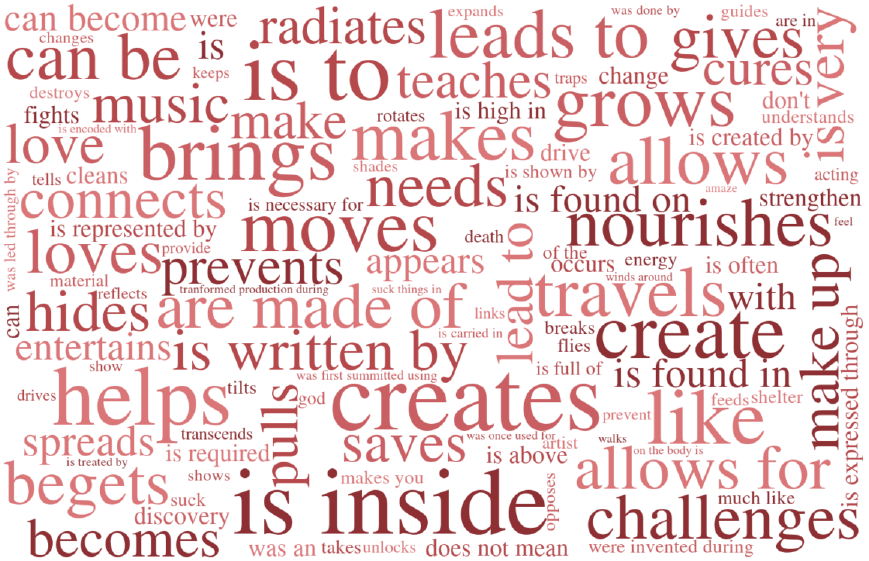}\caption{Human relations}\end{subfigure}\hfill
    \begin{subfigure}[t]{0.495\linewidth}\centering\includegraphics[width=\linewidth]{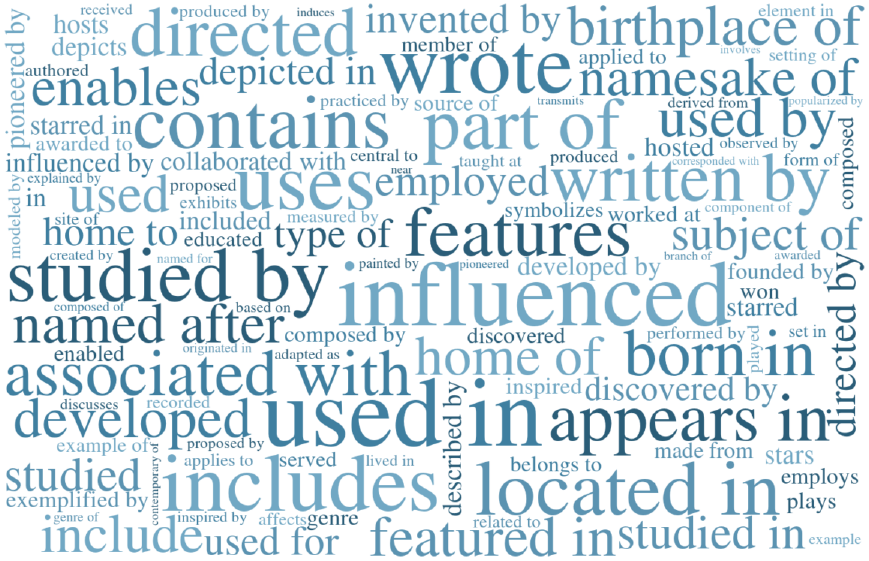}\caption{LLM relations}\end{subfigure}
    \caption{\textbf{Word cloud visualization of concepts and relations used by humans versus LLMs across all three tasks}. The size of each word corresponds to how much more often one population uses it than the other. \textbf{(a)} Human-generated concepts (entities) tend to be everyday objects, states, and feelings (e.g., ``rock,'' ``happiness,'' ``love''), while \textbf{(b)} LLM-generated concepts tend to be named people and places (e.g., ``James Clerk Maxwell,'' ``Galileo Galilei,'' ``London''). Similarly, \textbf{(c)} human-generated relations tend to express motion and plain states of being (e.g., ``leads to,'' ``moves,'' ``is inside''), while \textbf{(d)} LLM-generated relations tend to express creation, communication, and cognition (e.g., ``wrote,'' ``named after,'' ``studied by'').}
    \label{fig:human_llm_wordcloud}
\end{figure}

\subsection{Comparing Human and LLM Performance} \label{subsec:human_results}
In this section, we compare performance between the 35 LLMs and the 75 human participants described in \Cref{sec:benchmark}.

\textbf{Human versus LLM performance on each task} \quad Humans score below every LLM on the mean score across all three tasks (22.1\% vs.\ 59.0\% on association, 32.9\% vs.\ 53.4\% on analogy, and 12.8\% vs.\ 43.4\% on blending), as shown in \Cref{tab:leaderboard}. This difference is driven mainly by utility, since human participants were relatively poor at finding a valid connection on each task compared to LLMs. 

\textbf{Both humans and LLMs find blending most difficult, but association is easier than analogy for LLMs, while analogy is easier than association for humans} \quad While humans produce a valid analogy 32.9\% of the time and a factual association 22.1\% of the time, LLMs produce a valid analogy 53.4\% of the time, and a factual association 59.0\% of the time. Unlike LLMs, humans are able to find valid analogies more often than they can find valid associations. However, both populations find blending most difficult, with humans and LLMs only finding valid blends 12.8\% and 43.4\% of the time on average, respectively.

\begin{figure}[t]
\centering
\includegraphics[width=\linewidth]{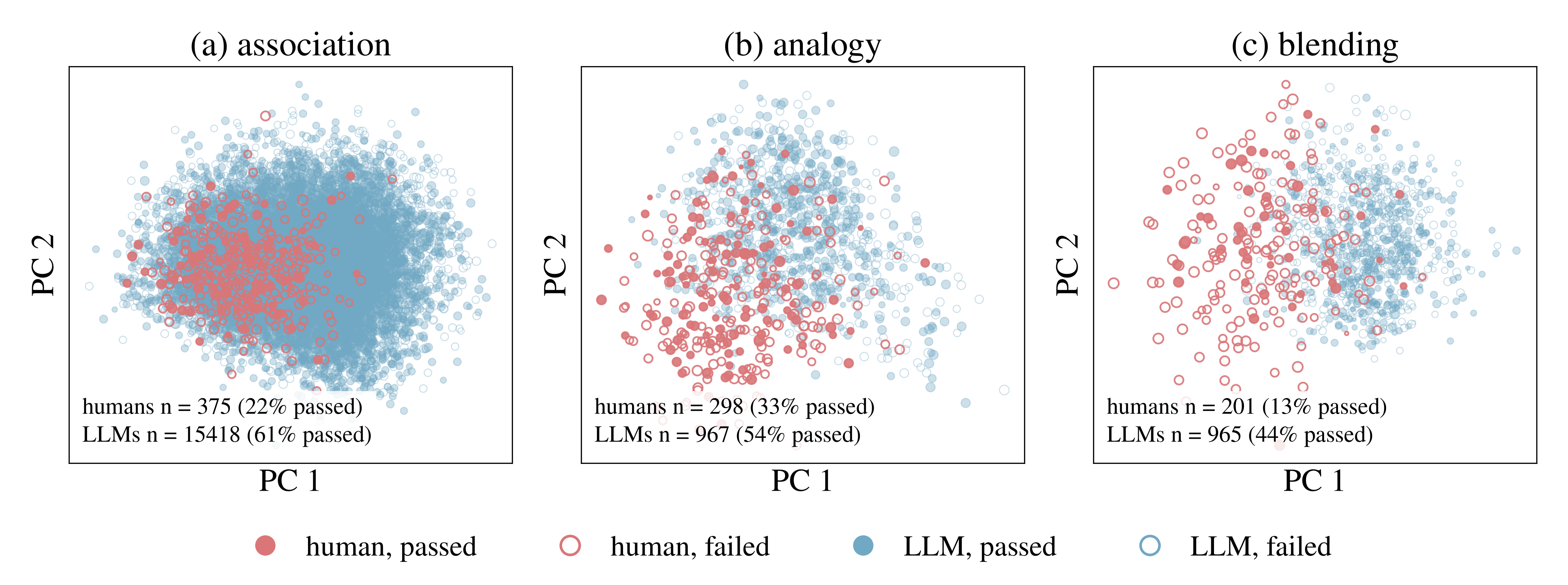}
\caption{\textbf{Visualization of human versus LLM responses in embedding space}. We embed the added elements of each response (every entity and relation besides the two supplied input entities) using \texttt{all-MiniLM-L6-v2}, subtract these from the mean response vector on each item, and plot the first two principal components of each response for \textbf{(a)} association, \textbf{(b)} analogy, and \textbf{(c)} blending. On analogy and blending in particular, humans and LLMs tend to occupy distinct semantic regions of the embedding space.}
\label{fig:human_llm_space}
\end{figure}

\textbf{Humans and LLMs access distinct regions of semantic space on each task} \quad \Cref{fig:human_llm_space} showcases human versus LLM responses in a two-dimensional embedding space, visualizing the distinct regions each population occupies on each task. Similarly, \Cref{fig:human_llm_wordcloud} portrays differences in entities and relations used between humans and LLMs. To characterize semantic differences between human and LLMs concepts and relations in a quantitative way, we label each concept with its WordNet \citep{miller1995wordnet} noun category and each relation with the category of its verb. Then, we compare how often human and LLM responses use each category on each task, and report frequency differences in percentage points (pp) between humans and LLMs. On association, LLMs more often name people (+5.2pp; \texttt{Emil Fischer, Galileo}) and places (+3.6pp; \texttt{London}), and connect them through creation (+4.3pp; \texttt{wrote, invented by}), communication (+3.8pp; \texttt{directed, named after}), and cognition (+2.5pp; \texttt{studied by, associated with}), whereas humans more often use physical objects (+3.0pp; \texttt{sun, rock}), states (+2.4pp; \texttt{disease, safety}), and attributes (+1.8pp; \texttt{power, sound}), and connect them through physical relations like motion (+7.6pp; \texttt{drives, leads to}). Similarly, on analogy humans more often use motion verbs (+6.9pp; \texttt{leads to, brings}), while LLMs more often change (+4.8pp; \texttt{enables, regulates, converts}), cognition (+3.6pp; \texttt{shapes, relies on}), and communication (+3.5pp; \texttt{determines, mediates}) verbs. Lastly, on blending, humans tend to connect concepts through plain ``stative'' relations (+7.6pp; \texttt{is, is a}), whereas LLMs use possession (+4.4pp; \texttt{distributes, transmits}), and LLM blends contain more concepts of communication (+6.3pp; \texttt{information, testimony}). In summary, LLMs tend to name people and places, and connect them through authorship, study and transformation (\texttt{wrote, studied by, converts}), while humans tend to use everyday objects, states, and feelings, and connect them through movement, being, and plain association (\texttt{leads to, is, like}).

% Human vs. LLM answers in meaning space (plot_human_vs_model_space.py, figure space_pca_utility, in the analysis repo):
% each answer is the mean MiniLM embedding of its concepts and relations, item-centred, then PCA per task;
% filled = passed utility, hollow = failed, size = originality. Caption left to the author.

\section{Discussion}

\paragraph{Implications of our findings on human-AI co-creativity}
Creativity is known to benefit from a diversity of perspectives \citep{simonton_discovery_invention_as_combinatorial, creativity_in_science, fang2025generalization, surprising_combinations}. While humans underperformed LLMs on each task, the two groups generated relations and entities from distinct semantic categories, suggesting that combinatorial creativity can benefit from human-AI collaboration to enhance conceptual and semantic diversity. Indeed, recent studies have begun to explore human-AI collaboration on creative tasks like the Alternative Uses Test \citep{hemmatian2026twoplayeralternateusestest, lai2026creative}, and extending this work to combinatorial creativity is a natural next step.

\paragraph{Comparing the creativity of humans and LLMs}
Our findings corroborate a broader trend among comparative studies of human and LLM intelligence finding that LLMs now tend to outperform humans on creative tasks. In particular, on an established psychometric test of divergent thinking known as the Divergent Association Test (DAT), LLMs now routinely exceed mean human scores \citep{schapiro2026assessingcreativitylargelanguage, bellemare2026divergent, Chen2023ProbingAssociation}. Nevertheless, LLMs are still afflicted by key weaknesses, such as the \emph{abstraction bottleneck} observed here (\Cref{subsec:llm_results}), or the \emph{ideation-execution gap} observed in \citet{Si2025TheIdeas, Si2024CanResearchers}, in which LLM-generated research ideas tend to be more novel than human ideas, but are less practically feasible. Even though LLMs now outperform humans in many ways, characterizing the (often ``jagged'') ways in which they fail remains a valuable research direction \citep{mineault2026cognitivedarkmattermeasuring}.

\paragraph{Why are analogies easier to make than blends?}
For both human and LLM populations, valid analogies were easier to find than valid blends. To understand why this might be the case, we draw upon the framework of \citet{elmoznino2025towards}, which formalizes compositionality as compression, such that a representation is compositional if it can be generated by a simple function of a short discrete description. Within this context, an analogy describes one input in terms of structure the other already has (roughly, compressing $v$ given $u$), whereas a generic space $g$ with structure both inputs $(u, v)$ instantiate must be constructed as a new description that jointly compresses both inputs. The latter may be more computationally demanding than compressing one input in terms of structure already available for another input.

\section{Limitations and Future Work} \label{sec:conclusion}
This work leaves several directions to future research. First, the human population recruited for this study is not representative of all human performance, especially scientists, artists, and creators more broadly, who may score higher on these tasks than the average participant. Extending this study to specific domains, such as combinatorial creativity within a particular scientific subfield, is a valuable direction for future work. Next, there is no broad consensus in the cognitive science literature on the degree of overlap between analogy and blending, and structure mapping theory is comprehensive enough to account for operations shared in both processes \citep{gentner_smith}. The core distinction is that the primary artifact of a blend is a blended space \citep{the_way_we_think}, whereas the outcome of analogy is typically the target domain projection of an element that already exists in the source domain \citep{petersen2026modelling}, although Figure 42.2(C) in \citet{gentner_smith} portrays a novel abstraction that can arguably be considered a type of blend. Therefore, analogy and blending should not be conceptualized as mutually exclusive forms (and, hence, are pictured as overlapping in \Cref{fig:generalization_taxonomy}), but rather as interdependent processes that co-occur empirically, with structure mapping involved in constructing a blend. Future work can explore the relationship between these abilities in greater detail. Similarly, although blending can involve more than two input spaces \citep{the_way_we_think}, we restrict to two in this work. Likewise, although humans employ combinatorial creativity to create visual art and music \citep{boden}, our benchmark is restricted to the textual domain. Lastly, \citet{boden} defines historical creativity (H-creativity) as novelty relative to human knowledge writ large, and personal creativity (P-creativity) as novelty relative only to an artifact's creator. Our benchmark randomly samples knowledge graph entities for which prior relationships are not established, and thus approaches H-creativity, but our originality measures cannot guarantee the historical novelty of all artifacts created. Therefore, our benchmark assesses a level of creativity situated between P- and H-creativity. 

\section{Conclusion}
In this work, we introduced \textsc{Kombine}, a benchmark of combinatorial creativity for LLMs, assessing their ability to discover associations, analogies, and blends between distant, unrelated entities. We found that models were 10-16 percentage points better at discovering associations and analogies than blends, and that the main bottleneck on blending was abstraction ability, a weakness we refer to as \emph{the abstraction bottleneck}. We also found that when tasked to invent new concepts, LLMs often reused properties invented by other models, in what we refer to as \emph{inventive homogeneity}. This homogeneity was more pronounced among models from the same provider, as well as on the blending task versus the analogy task. Lastly, although humans scored below nearly every LLM on all three tasks, we found that humans and LLMs tend to access distinct regions of semantic space on each task, suggesting that combinatorial creativity could nevertheless benefit from systematic human-AI collaboration to enhance conceptual diversity.
\section*{Reproducibility Statement}
To ensure reproducibility of results, the entities and procedure used to construct the benchmark are fully listed in \Cref{app:entity_pool}, the task prompts are given in \Cref{app:prompts}, and the judge prompts are given in \Cref{app:judge_prompts}. Every model is elicited under a recorded configuration giving the model identifier and the sampling temperature ($T = 0.9$, with top-$p$ and top-$k$ disabled). Embedding-based scoring dimensions use \texttt{all-MiniLM-L6-v2}, although the leaderboard is highly robust to the choice of embedding model, as shown in \Cref{app:embedding_robustness}.

\section*{Acknowledgements}
We thank Toyota Motor North America, Inc. for their support of this research.

A.G. is supported by the National Science Foundation Graduate Research Fellowship Program under Grant No. DGE 21-46756. Any opinions, findings, and conclusions or recommendations expressed in this material are those of the author(s) and do not necessarily reflect the views of the National Science Foundation. 

A.K.J. is supported by a Natural and Artificial Mind (NAM) Fellowship from the Scully Peretsman foundation
\section*{Author Contributions}
S. Sc. designed all tasks, conducted LLM and human experiments, and wrote the majority of the paper; A.K.J. assisted with human experiments, task design, and writing; A. G. assisted with LLM experiments and provided feedback on writing; Q.E.L. assisted with task design, writing, and literature review; S. Sh. assisted with LLM experiments; J.B. assisted with writing and task design; R. M. assisted with writing and figure design; D. H-T. assisted with LLM experiments; C.E.S. assisted with writing and task design; L.R.V. assisted with writing and task design; R.E.B. assisted with writing and task design, and acquired funding for LLM experiments; T.L.G. assisted with task design, human experiments, and acquired funding for human experiments; J.A.E. assisted with writing, task design, and figures.

\bibliography{setup/iclr}

\newpage
\appendix
\section{Additional Background} \label{app:additional_creativity_background}

\paragraph{Exploratory and Transformational Creativity} 
Exploratory and transformational creativity are historically defined with respect to a  ``conceptual space,'' a set of rules and constraints that defines what constitutes well-defined and intelligible artifacts in a particular domain. Exploratory creativity refers to artifacts generated by following these rules and constraints (such as AlphaGo move 37 \citep{alphago}) whereas transformational creativity, which refers to the more challenging task of re-structuring the very rules of a conceptual space, is considered the pinnacle form of creativity for its historical role in breakthrough innovation \citep{boden}. Famous examples of transformational creativity include Einstein's relativity theory, the shift from geocentrism to heliocentrism, and the invention of non-Euclidean geometry \citep{haven100,trans_graphical_theory,conceptual_revolutions, koestler_creation}, though transformational creativity is also implicated in other creative areas---such as in the literary domain when producing plot twists \citep{schapiro2026twistbench}.

\paragraph{Divergent and Convergent Thinking}
Another popular taxonomy of creative processes distinguishes between divergent and convergent thinking \citep{Dietrich2019TypesCreativity, Dietrich2004TheCreativity}. Broadly speaking, divergent thinking refers to the ability to generate multiple diverse stimuli on an open-ended task, while convergent thinking is the ability to generate a single response that unifies multiple diverse stimuli. Divergent and convergent creativity are commonly assessed using psychometric tests such as the Divergent Association Task \citep{Olson2021NamingCreativity}, Remote Associates Test \citep{associative_basis}, and more recently, Parallel Associative Chains for Creativity Evaluation \citep{PACE}, the Conditional Divergent Association Task \citep{nakajima2026divergentcreativityhumanbasedevaluation}, and the Divergent Remote Association Task \citep{schapiro2026assessingcreativitylargelanguage}. However, as shown in \citet{schapiro2026assessingcreativitylargelanguage}, psychometric tests designed for human populations often show poor external validity when applied to LLMs, casting doubt on their efficacy as instruments for studying LLM intelligence.
\section{Entity Pool} \label{app:entity_pool}
As shown in \Cref{tab:entity_pool}, the space of entities $\mathcal{C}$ consists of 283 entities spread across 21 domains. Entities were filtered for their recognizability, and each entity has a median of 164 Wikimedia sitelinks, with a minimum of 78 across all entities. In contrast, entities drawn uniformly at random from Wikidata have a \emph{median of zero} sitelinks and a $90^{\text{th}}$ percentile of $2$. Furthermore, domains are balanced by construction, at 10--16 entities each, and the entity input pairs for each of the three tasks consists of pairs from different domains. 

\begin{table}[tbp]
\centering
\caption{\textbf{The entity pool $\mathcal{C}$.} All 283 entities, categorized by domain.}
\label{tab:entity_pool}
\small
\renewcommand{\arraystretch}{1.15}
\begin{tabularx}{\textwidth}{@{}lX@{}}
\toprule
\textbf{Domain} & \textbf{Entities} \\
\midrule
\textbf{science} & Gravity, Evolution, Entropy, Photosynthesis, Electricity, Radioactivity, The atom, Quantum mechanics, The theory of relativity, Plate tectonics, The Big Bang, Natural selection, The periodic table, Nuclear fission, Magnetism, The speed of light \\
\textbf{mathematics} & Zero, Infinity, Pi, Prime numbers, The Pythagorean theorem, Calculus, Probability, Geometry, The golden ratio, Fractals, Chaos theory, Set theory, Logarithms, Fibonacci sequence \\
\textbf{technology} & The printing press, The steam engine, The internet, The telephone, The computer, The transistor, The automobile, The airplane, The light bulb, The wheel, Artificial intelligence, The battery, Radio, GPS, The camera, The clock \\
\textbf{biology} & DNA, The cell, The brain, The heart, Bacteria, Viruses, The immune system, Photosynthesis, The human genome, Coral reef, The honey bee, The lion, The octopus, The oak tree, Mushrooms, The blue whale \\
\textbf{medicine} & Penicillin, Vaccines, Anesthesia, The stethoscope, Antibiotics, Surgery, X-rays, Insulin, The Hippocratic Oath, Blood transfusion, Aspirin, The microscope \\
\textbf{art} & The Mona Lisa, Impressionism, Cubism, Leonardo da Vinci, Pablo Picasso, Vincent van Gogh, Frida Kahlo, Michelangelo, The Sistine Chapel, Surrealism, Sculpture, Andy Warhol, Calligraphy, Origami \\
\textbf{music} & Jazz, The symphony, The piano, Ludwig van Beethoven, The Beatles, Wolfgang Amadeus Mozart, Hip hop, The guitar, Opera, Bob Dylan, The blues, Reggae, The orchestra, Rock and roll \\
\textbf{literature} & William Shakespeare, The novel, Poetry, Homer's Odyssey, Don Quixote, Jane Austen, Greek tragedy, Haiku, Leo Tolstoy, Franz Kafka, The sonnet, Mythology, Science fiction, Fairy tales \\
\textbf{film} & Cinema, Alfred Hitchcock, Charlie Chaplin, Animation, Akira Kurosawa, Documentary film, Steven Spielberg, Film noir, Walt Disney, The Western \\
\textbf{philosophy} & Democracy, Free will, Justice, Aristotle, Plato, Socrates, Existentialism, Ethics, Immanuel Kant, Stoicism, Confucius, The social contract, Logic, Nihilism \\
\textbf{religion} & Buddhism, Christianity, Islam, Hinduism, Judaism, Meditation, The Ten Commandments, Karma, Pilgrimage, Zen, Prayer, Mythology \\
\textbf{history} & The Roman Empire, The Renaissance, The French Revolution, The Industrial Revolution, Ancient Egypt, The Silk Road, World War II, The Cold War, The Enlightenment, Feudalism, The Great Depression, Colonialism, The Space Race \\
\textbf{geography} & Mount Everest, The Amazon rainforest, The Sahara, The Nile, The Great Barrier Reef, Antarctica, The Grand Canyon, The Himalayas, Volcanoes, The Pacific Ocean, Iceland, Japan, The Alps, Niagara Falls \\
\textbf{politics} & Abraham Lincoln, Nelson Mandela, Mahatma Gandhi, Napoleon, Winston Churchill, The United Nations, Cleopatra, Julius Caesar, The Constitution, Martin Luther King Jr., Diplomacy, The monarchy \\
\textbf{economics} & Money, The stock market, Capitalism, Supply and demand, Adam Smith, Inflation, Banking, Trade, Cryptocurrency, The Great Depression, Taxation, Globalization \\
\textbf{sports} & Chess, Football (soccer), The Olympic Games, Boxing, Muhammad Ali, Basketball, Marathon running, Tennis, Gymnastics, Surfing, Cricket, Formula One \\
\textbf{food} & Coffee, Chocolate, Wine, Bread, Rice, Tea, Cheese, Sushi, Pizza, Chili peppers, Salt, Honey, Fermentation, Spices \\
\textbf{everyday} & The mirror, The lock and key, Glass, Paper, The umbrella, Scissors, Soap, The mattress, Candles, The map, The button, Rope, The needle, The compass \\
\textbf{nature} & Fire, Lightning, The rainbow, Snow, The tides, Earthquakes, Diamonds, Gold, Rivers, Deserts, The seasons, Fog, Amber, Crystals \\
\textbf{cosmos} & The Sun, The Moon, Mars, The solar system, Black holes, The Milky Way, Comets, The Northern Lights, Saturn's rings, Eclipses, Meteorites, Constellations \\
\textbf{ideas} & Time, Language, Color, Symmetry, Chaos, Memory, Consciousness, Beauty, Randomness, Emergence, Feedback loops, Metaphor, Paradox, Networks \\
\bottomrule
\end{tabularx}
\end{table}

\clearpage
\section{Full Task Prompts} \label{app:prompts}
The complete prompt for each of the three tasks is given below.  Placeholders \texttt{\{u\}} and \texttt{\{v\}} denote the randomly sampled anchor entities, following the procedure in \Cref{app:entity_pool}.
\begin{promptbox}[Task and Prompt: \emph{Association}]
Query: What are different ways in which '{u}' is connected to '{v}'?

Task: Identify how these two entities are connected by producing as MANY DISTINCT connection paths as you can. A connection path is a sequence of factual triples (head, relationship, tail) forming a continuous chain: consecutive triples share an entity, the first triple's head is '{u}', and the last triple's tail is '{v}'.

We reward three things in every connection:
- TRUE: every triple is factually correct.
- REMOTE: the intermediate concepts sit in domains far from the two endpoints and from each other -- reach across distant fields rather than taking the first obvious link.
- UNCOMMON: build the path from rare, specific concepts and relations, not generic ones most people would give.

Produce as MANY DISTINCT connections as you can, most surprising first -- do not stop at a fixed number; list every genuine connection you can find, and make them as different from one another as possible (different intermediate entities and relations, spanning different domains).


Rules and quality constraints:
- Entities may be concrete or abstract (people, works, places, species, events, ideas, phenomena, theories, etc.), but must be real and recognizable -- do not invent entities.
- Do not ask follow-up questions; respond using the best available factual knowledge.
- Temporal connections are allowed (relationships may span different historical periods).
- Disambiguation is required: use canonical names and qualifiers where necessary (e.g., 'Michael Jordan (basketball)').

Deduplication:
- Do not repeat the same path, and do not repeat the same entity within a single path.
- Prefer paths that are meaningfully different (different intermediate nodes and/or relationships), not trivial rephrasings.

Output requirements (strict):
- Return ONLY a JSON array wrapped in <answer> and </answer> tags. No other text, before or after.
- Each element of the array is ONE connection: an object with a single key, "path".
- "path" is a list of triples; each triple is [head entity, relationship, tail entity] forming a continuous chain from the first entity to the last.
- Relationship strings must be 1-3 words. Keep every entity SHORT and CONCRETE: a few plain, recognizable words naming a real concept -- not a descriptive phrase, a clause, an invented CamelCase compound, or words joined by a dash into a coinage (write "petition threshold", never "PetitionThreshold" or "petition-threshold").
- List one object per connection. To stop, simply end the array -- do not pad with weak connections.
- If you can find no valid connection, return an empty array: <answer>[]</answer>.

Required format (follow this shape exactly):
<answer>[{"path": [["A", "r1", "B"], ["B", "r2", "C"]]}, {"path": [["A", "s1", "D"], ["D", "s2", "C"]]}]</answer>
\end{promptbox}
\captionof{figure}{\textbf{Association task prompt.} The full instructions administered for the association task; \texttt{\{u\}} and \texttt{\{v\}} are the two randomly sampled anchor entities.}
\label{fig:prompt_association}

\newpage
\begin{promptbox}[Task and Prompt: \emph{Analogy}]
Task: You are given two concepts: '{u}' and '{v}'. Build the single best analogy between them, and use it to INVENT a new concept in one of the two domains.

An analogy aligns the two domains through a shared relational structure -- '{u}' plays a role in its domain analogous to the role '{v}' plays in its. You then use that alignment to invent something new: carry a concept from one domain across the mapping into the other.

We reward four things, matching exactly how the analogy is scored:
- UTILITY: every triple in "path_a" and "path_b" is factually correct, and the two paths share an IDENTICAL relation sequence -- a genuine structural correspondence, not a loose resemblance.
- SURPRISE: the aligned entities in corresponding roles of the two paths are as semantically DISTANT as possible, so the mapping travels far across domains.
- ORIGINALITY: build the analogy from RARE, uncommon entities and relations, far from the obvious ones others would give.
- EMERGENT CREATIVITY (the invention): pick a real "projected" concept from ONE domain that has NO counterpart in the other, and carry its structure across the mapping to INVENT a new concept in the other domain -- rewarded for being novel, internally coherent, and a genuine projection through the mapping (the invention need NOT already exist -- inventing it is the point, as the solar system did for the atom).

Each analogy is TWO paths that share an identical relation sequence:
- "path_a": factual triples describing '{u}' within its own domain, beginning at '{u}'.
- "path_b": factual triples describing '{v}' within its own domain, beginning at '{v}'.
CRITICAL: "path_a" and "path_b" must use the EXACT SAME relationship word at every position (only the entities differ), so that position i of "path_a" corresponds to position i of "path_b". Do NOT paraphrase or substitute synonyms (e.g. if "path_a" uses 'awards', "path_b" must also use 'awards', not 'grants'). Use disjoint, recognizable, canonically-named entities, and do not repeat an entity within a path.

Produce exactly ONE analogy -- the single most surprising, genuine analogy you can build, together with its invention.

Output requirements (strict):
- Return ONLY a JSON array wrapped in <answer> and </answer> tags, containing EXACTLY ONE analogy object. No other text, before or after.
- The single array element is the analogy: an object with keys "path_a", "path_b", "projected", "invention", and "projection".
- "path_a" and "path_b" are each a list of triples [head, relationship, tail] that establish the mapping. They MUST have the same number of triples and the IDENTICAL relationship word at every position (only the entities differ); the entity at position i of "path_a" corresponds to the entity at position i of "path_b".
- "projected" is a real concept from ONE domain (the source) that has NO counterpart in the other -- the thing you carry across the mapping to INVENT something new (e.g. "vaccine").
- "invention" is a short name for the NEW concept the projection creates in the other (target) domain.
- "projection" is a list of {"source": [triple], "image": [triple]} pairs. Each "source" is a true triple about the "projected" concept in its own domain; each "image" carries it across the mapping onto the invention -- same relationship, entities replaced by their counterparts. The invention need NOT already exist (that is the point), but every "image" must be a coherent projection of its "source".
- Relationship strings must be 1-3 words. Keep every entity SHORT and CONCRETE: a few plain, recognizable words naming a real concept -- not a descriptive phrase, a clause, an invented CamelCase compound, or words joined by a dash into a coinage (write "petition threshold", never "PetitionThreshold" or "petition-threshold").
- The array must contain exactly one analogy object.
- If you can find no valid analogy, return an empty array: <answer>[]</answer>.

Required format (follow this shape exactly):
<answer>[{"path_a": [["A", "r1", "B"], ["B", "r2", "C"]], "path_b": [["D", "r1", "E"], ["E", "r2", "F"]], "projected": "X", "invention": "Y", "projection": [{"source": ["X", "r", "B"], "image": ["Y", "r", "E"]}]}]</answer>
\end{promptbox}
\captionof{figure}{\textbf{Analogy task prompt.} The full instructions administered for the analogy task; \texttt{\{u\}} and \texttt{\{v\}} are the two randomly sampled anchor entities.}
\label{fig:prompt_analogy}

\newpage
\begin{promptbox}[Task and Prompt: \emph{Blending}]
Task: You are given TWO concepts: '{u}' and '{v}'. FUSE them into a SINGLE new concept, then describe the structure this fusion generates.

What a genuine blend is (the FORM). A real blend fuses two concepts into ONE new concept in which BOTH inputs contribute ORGANIZING STRUCTURE to the SAME property -- one slot of the blend is fed by both inputs at once, so their frames combine there rather than sit side by side. The classic case is a "liquid franchise" (democracy + banking): its one "allocates" slot carries BOTH democracy's "allocates votes" AND banking's "allocates credit", fusing into "allocates transferable vote-shares" -- a single property both inputs organize.
Example -- "cyborg" (organism + machine): the organism frame (a living system that self-regulates through homeostasis) AND the machine frame (a control system that self-regulates through feedback) BOTH organize its ONE "regulates itself" slot; and it has emergent structure of neither input: its components can be surgically replaced without the downtime a body needs to let tissue heal -- neither a pure organism (which must heal) nor a pure machine (which has no tissue) works this way.
(Note what does NOT count: "it can be rebooted" is already true of the machine alone and "it can die" of the organism alone, so those are inherited, not emergent; and listing "can die" beside "has components" is only a concatenation until some ONE property is fed by both.)
Do NOT instead (a) list one property from each input side by side with no shared slot ("harnesses energy" from one AND "is therapeutic" from the other), or (b) treat one input as a mere adjective on the other ("a radioactive solar system"). At least one property must be organized by BOTH inputs at once.

Build the blend in two moves:
1. GENERIC SPACE: name the shared schema both '{u}' and '{v}' instantiate -- the abstract structure that lets them fuse (for organism + machine: "a functional system whose parts can fail and be replaced"). Be specific; "both exist"/"both involve change" is vacuous, and a one-from-each conjunction does not count.
2. STRUCTURE: describe the single blended concept as triples, and TAG each triple by where its structure comes from -- "u" (organized by '{u}' only), "v" (organized by '{v}' only), "uv" (ONE property organized by BOTH inputs at once -- the same slot carries both frames, like the liquid franchise's single "allocates" slot fed by votes AND credit), or "emergent" (true of the BLEND but of NEITHER '{u}' nor '{v}' alone -- structure that appears only when you RUN the fusion forward). A genuine blend has at least one "uv" triple -- a property both inputs feed -- plus at least one "emergent"; separate "u" and "v" lines that never share a slot are a concatenation, not a fusion.

We reward three things:
- UTILITY: the generic space is a REAL, specific schema that BOTH '{u}' and '{v}' genuinely instantiate -- not vacuous ("both exist") and not a one-from-each conjunction.
- ORIGINALITY: the generic space is UNCOMMON -- a shared schema few would think to name, not the first obvious one.
- EMERGENT CREATIVITY: a genuine DOUBLE-SCOPE fusion in which at least one property is organized by BOTH inputs at once (tag it "uv"), developed into a coherent, original blended concept with EMERGENT structure true of the BLEND but of NEITHER '{u}' alone NOR '{v}' alone (tag these "emergent"). Merely listing one input's properties beside the other's, with no shared slot, does not count.

Produce exactly ONE blend of '{u}' and '{v}'. Keep the structure tight: give the 4-6 triples that best showcase the blend's most important properties (at least one "uv" property both inputs organize, plus at least one emergent), not an exhaustive dump. Use recognizable, canonically-named entities in the structure.

Output requirements (strict):
- Return ONLY a SINGLE JSON object wrapped in <answer> and </answer> tags. No other text, before or after. (Produce ONE blend, not a list.)
- The object has exactly these three keys: "concept", "generic_space", "structure".
- "concept": a short name for the single blended concept you create by fusing the two inputs.
- "generic_space": ONE phrase naming the shared schema both inputs fit -- what makes them fusable. Be specific; a vacuous schema ("both exist", "both involve change") does not count as a blend.
- "structure": a list of objects {"triple": [head, relationship, tail], "from": TAG} describing the blend, where TAG is EXACTLY one of: "u" (this property is organized by the FIRST input only), "v" (this property is organized by the SECOND input only), "uv" (a SINGLE property that BOTH inputs organize at once -- the same slot carries both frames, e.g. a "liquid franchise" whose one "allocates" slot is fed by democracy's "allocates votes" AND banking's "allocates credit"), or "emergent" (this property is true of the BLEND but of NEITHER input alone; it arises only from running the fusion). A genuine blend has at least one "uv" triple -- a property both inputs feed -- plus at least one "emergent"; separate "u" and "v" lines that never share a slot are a concatenation, not a fusion.
- Relationship strings must be 1-3 words; the head is usually the blend. Keep every entity SHORT and CONCRETE: a few plain, recognizable words naming a real concept -- not a descriptive phrase, a clause, an invented CamelCase compound, or words joined by a dash into a coinage (write "voting claims", never "voting-claims"). State the generic space briefly and in plain language too, even when the underlying schema is abstract.

Required format (follow this shape exactly):
<answer>{"concept": "cyborg", "generic_space": "a functional system whose parts can fail and be replaced", "structure": [{"triple": ["cyborg", "can", "die"], "from": "u"}, {"triple": ["cyborg", "has", "components"], "from": "v"}, {"triple": ["cyborg", "regulates itself via", "feedback"], "from": "uv"}, {"triple": ["cyborg", "replaces components", "without healing downtime"], "from": "emergent"}]}</answer>
\end{promptbox}
\captionof{figure}{\textbf{Blending task prompt.} The full instructions administered for the blending task; \texttt{\{u\}} and \texttt{\{v\}} are the two randomly sampled anchor entities.}
\label{fig:prompt_blending}

\clearpage
\section{Human Study Platform} \label{app:human_platform}

\begin{figure}[H]
\centering
\includegraphics[width=0.75\linewidth]{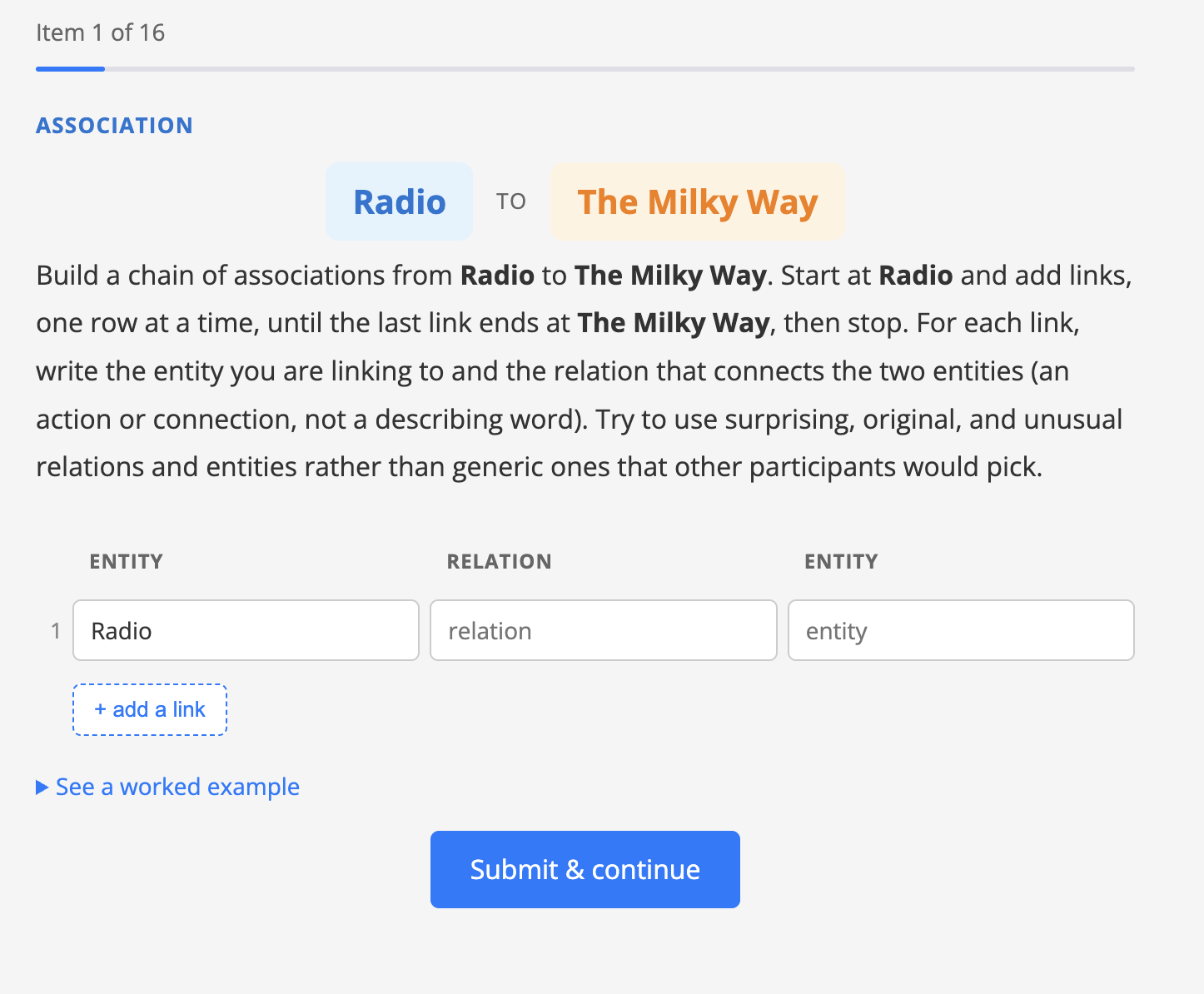}
\caption{\textbf{Association task interface.} Participants are instructed to build a path from $u$ to $v$ one link at a time, adding entity-relation-entity triples until the last entity is $v$.}
\label{fig:platform_association}
\end{figure}

\begin{figure}[H]
\centering
\includegraphics[width=\linewidth]{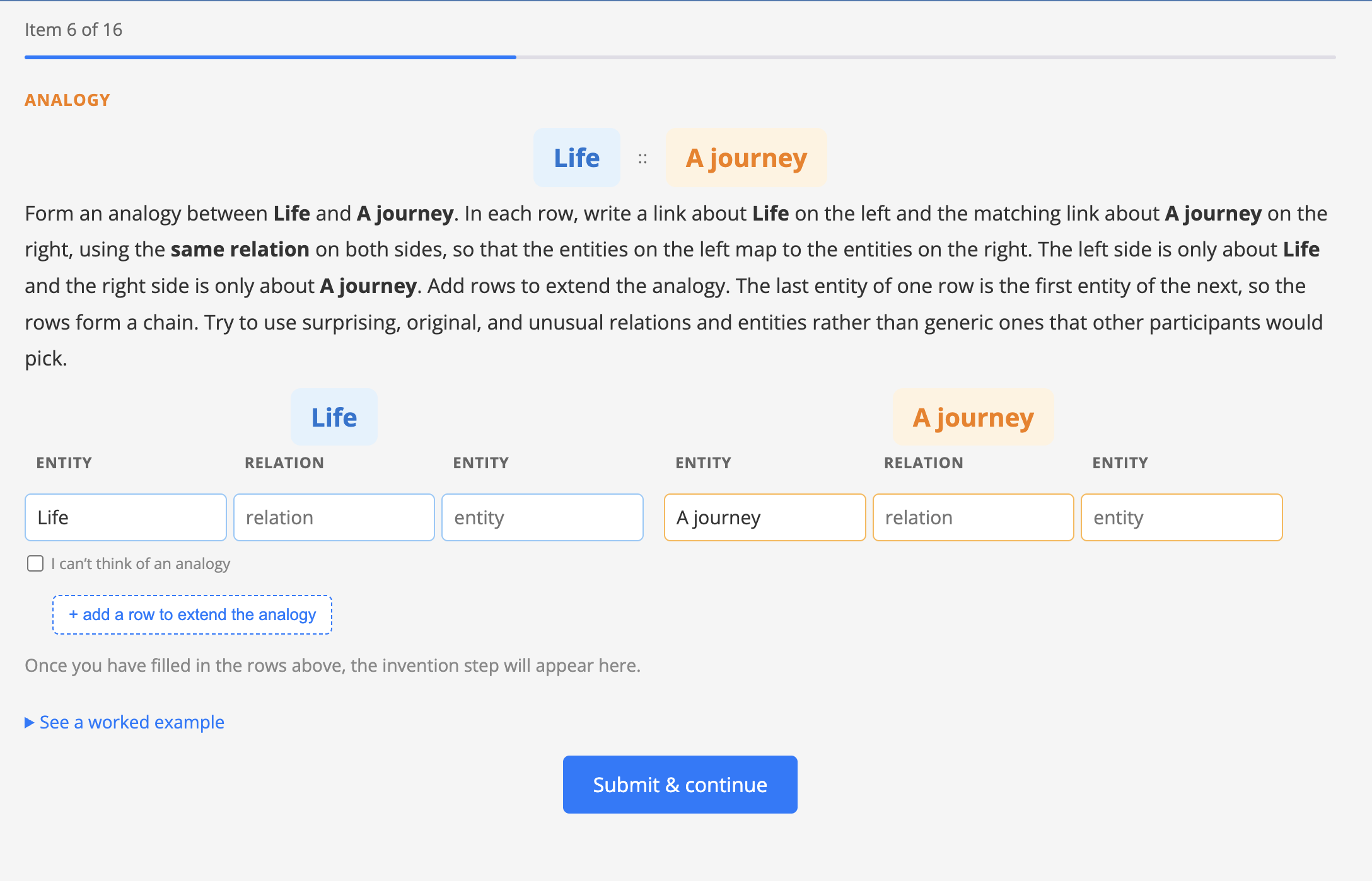}
\caption{\textbf{Analogy task interface,} shown on the \texttt{Life :: A journey} warm-up item. The invention step appears once the rows of the analogy are filled in.}
\label{fig:platform_analogy}
\end{figure}

\begin{figure}[H]
\centering
\includegraphics[width=0.9\linewidth]{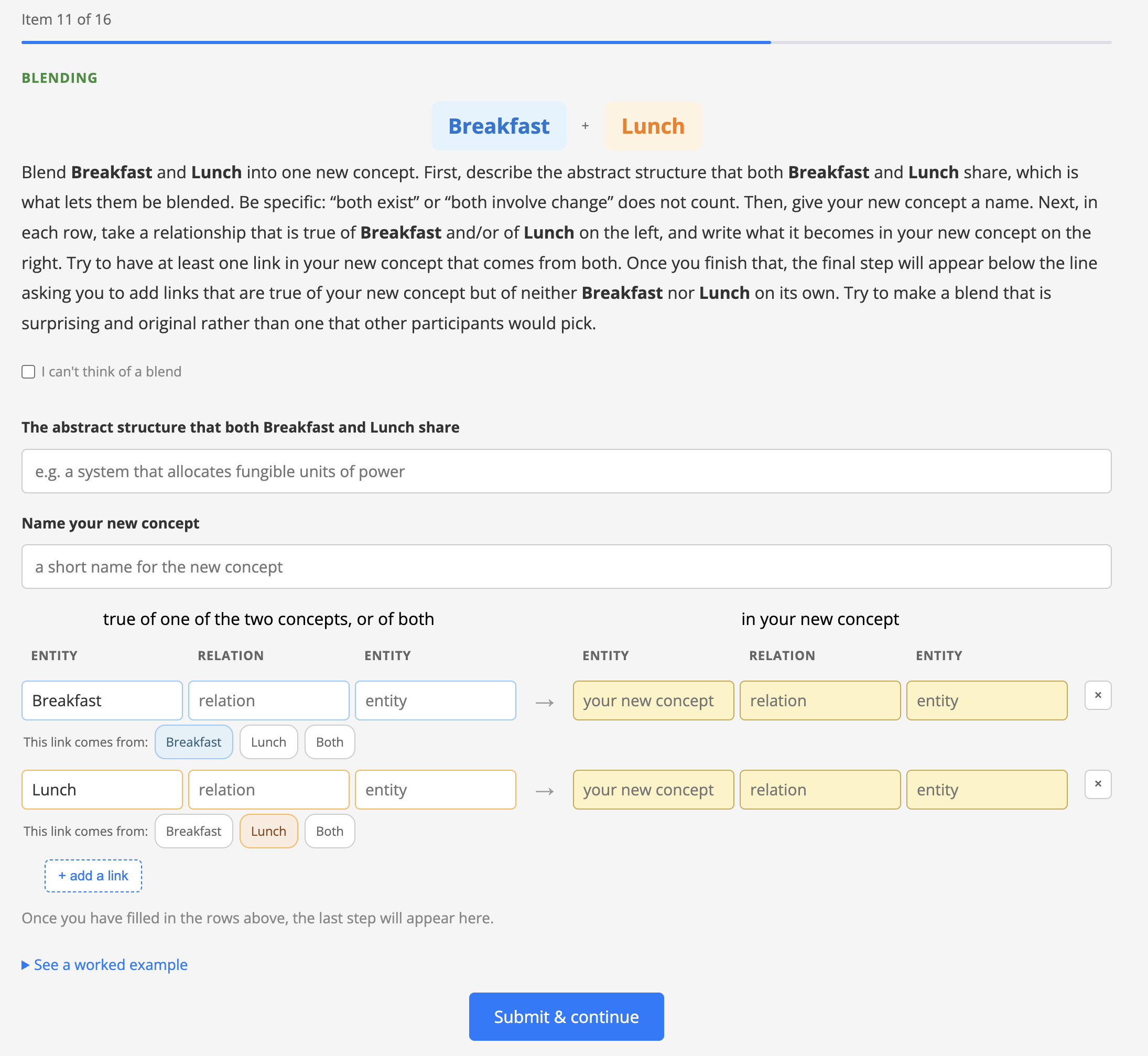}
\caption{\textbf{Blending task interface,} shown on the \texttt{Breakfast + Lunch} warm-up item. The final step, which asks for emergent properties true of the blended concept but of neither input, appears once the rows are filled in.}
\label{fig:platform_blending}
\end{figure}

\Cref{fig:platform_association,fig:platform_analogy,fig:platform_blending} show the web interface participants used for each of the three tasks, with the instructions shown on every item.

\clearpage
\section{Judge Prompts} \label{app:judge_prompts}
The judges prompts used in \Cref{sec:benchmark} are given below. The judges consist of a factuality judge $F$ used for determining utility on association and analogy, as well as the emergent creativity judges for analogy (coherence and validity of the invention) and blending (generic-space utility, coherence, and the double-scope quality score $Q_\text{bl}$).
\begin{promptbox}[Judge Prompt: \emph{Factuality} ($F$)]
You are a fact-checking expert evaluating a multi-hop factual path for factual accuracy and logical validity. The path consists of an ordered list of triples. Each triple has the form: (subject, relation, object). Your task is to evaluate EACH triple in the given path and determine whether it is hallucinated or not. Use the path only for entity disambiguation and contextual grounding. Do not infer unsupported facts.

Evaluation criteria (apply independently to each triple):
1. Entity existence and disambiguation: verify that both the subject and object are real, identifiable entities (people, places, organizations, events, or concepts). Use the path context only when necessary to disambiguate entities. If either entity is fabricated or incorrectly identified, mark the triple as hallucinated.
2. Relationship validity: verify that the stated relation describes a real, historically valid, currently valid, or plausibly valid relationship between the subject and object. If the relationship is plausible and could reasonably have occurred, even if not directly verifiable, mark it as not hallucinated. If it is niche information that only trivia experts would know, do not mark it as hallucinated.
3. Hallucination detection: mark the triple as hallucinated if an entity is fabricated or does not exist; the relation is fabricated, nonsensical, or incorrect; the asserted connection clearly contradicts well-known facts; or the triple introduces made-up properties or roles.
4. Directionality: if the relation is ambiguous or commonly bidirectional, evaluate both directions; if either direction corresponds to a true or plausible relationship, mark the triple as not hallucinated.

Output instructions: return valid JSON only, with exactly two keys, "explanation" and "judgments". Under "judgments", give the final judgment for each triple as a list, in the same order as the path. Do not include any text outside the JSON object.

The output must exactly match this schema:
{ "explanation": "string", "judgments": ["hallucinated" | "not hallucinated", ...] }

Input:
Path: {path}
\end{promptbox}
\captionof{figure}{\textbf{Factuality judge.} We reuse the per-triple factuality judge prompt from \citet{wadhwa2026createtestingllmsassociative} to assess \emph{utility} for association and analogy.}
\label{fig:judge_factuality}

\newpage
\begin{promptbox}[Judge Prompt: \emph{Analogy invention} ($Q(h)$, $V(h)$)]
You are judging an ANALOGY that invents a new concept by projecting structure across a mapping.

You are given two aligned paths (position i of path_a corresponds to position i of path_b), a "projected" source concept, an "invention" name, and a "projection": a list of (source, image) triple pairs.
path_a: {path_a}
path_b: {path_b}
projected: {projected}
invention: {invention}
projection: {projection}

Decide two things independently:
1. INTEGRATION QUALITY (valid): was the mapping ACTUALLY used to invent the concept -- i.e., is the invention the source structure carried faithfully across the mapping? Both must hold: (a) each "source" triple is factually true of the "projected" concept in its own domain, so a real source structure is being projected; (b) each "image" is that "source" carried across the mapping -- same relationship, every entity replaced by its counterpart under the path alignment (path_a[i] corresponds to path_b[i]), with any newly introduced counterparts used consistently. Judge ONLY whether the projection was genuinely applied; do not reward or penalize how novel or unusual the invention is (novelty is scored separately).
2. UTILITY (coherent): is the invention feasible and coherent -- its "image" triples each individually plausible or sensible, and together describing a single usable concept rather than a disjointed list? The invention need not be a real or existing thing.

Return valid JSON only, exactly:
{ "explanation": "string", "valid": true or false, "coherent": true or false }
\end{promptbox}
\captionof{figure}{\textbf{Analogy invention judge.} This judge determines the invention's integration quality ($Q(h)$: was the mapping actually used?) and utility ($V(h)$: is the invention feasible and coherent as one concept?).}
\label{fig:judge_analogy}

\newpage
\begin{promptbox}[Judge Prompt: \emph{Blend} ($G(g)$, $V(c')$, $Q(c')$)]
You are judging a BLEND that fuses two input concepts into a single new concept. A genuine blend has BOTH inputs contribute organizing structure to the SAME property -- one slot fed by both at once (e.g. a "liquid franchise" whose single "allocates" slot carries BOTH democracy's "allocates votes" AND banking's "allocates credit"), not two lists of properties placed side by side.

Input 1 (u): {u}
Input 2 (v): {v}
Blend concept: {concept}
Generic space (claimed shared schema): {generic_space}
Structure (each triple tagged "u" = from input 1 only, "v" = from input 2 only, "uv" = a single property BOTH inputs organize, or "emergent" = true of the blend but of neither input alone):
{structure}

Decide the following independently:
1. GENERIC SPACE (generic_ok): is the generic space a REAL, specific schema that BOTH '{u}' and '{v}' genuinely instantiate -- not vacuous ("both exist", "both involve change"), and not a one-from-each conjunction? (true/false)
2. UTILITY (coherent): are the triples in the blended space feasible or coherent -- each individually plausible, and together describing a single coherent new concept rather than two lists of properties placed side by side? The blend need not be a real or existing thing. (true/false)
3. TAGS: verify each triple's tag by content, not label. A "u"/"v" triple must be organizing structure genuinely from that ONE input; a "uv" triple must be a SINGLE property BOTH inputs genuinely organize (both frames on the same slot); an "emergent" triple must be true of the BLEND but of NEITHER '{u}' alone NOR '{v}' alone (a property already true of one input is inherited, not emergent). Ignore a triple whose tag is wrong, but credit a genuine shared slot even if mis-tagged.
4. SCOPE (scope): using only the correctly-tagged structure, classify the fusion:
   - 1 (single-scope): NO genuine shared slot -- no single property is organized by both inputs;
   - 2 (double-scope): at least one genuine shared slot -- a property BOTH inputs organize at once, a real "uv";
   - 3 (double-scope emergent): a genuine shared slot AND at least one genuine "emergent" triple.

Return valid JSON only, exactly:
{ "explanation": "string", "generic_ok": true or false, "coherent": true or false, "scope": 1 or 2 or 3, "shared_properties": ["name each property both inputs organize"] }
\end{promptbox}
\captionof{figure}{\textbf{Blend judge.} This judge scores the blend's \emph{utility} ($G(g)$: is the generic space a real shared schema both inputs instantiate?) and two emergent creativity dimensions: the invention's utility ($V(c')$: is the blend feasible and coherent?) and the semantic integration quality ($Q(c')\in\{1,2,3\}$).}
\label{fig:judge_blend}

\newpage
\section{Judge Reliability and Human Corroboration} \label{app:reliability}

The subjectively scored dimensions (analogy emergent utility, analogy emergent semantic integration, blend utility, blend emergent utility, and blend emergent semantic integration) are scored by a panel of three LLM judges (Claude Haiku 4.5, GPT-5.4, and GPT-o3) and aggregated by their mean. \Cref{tab:icc} reports the intra-class correlation coefficient \citep{shrout1979intraclass} for each subjective dimension. The intra-class correlation coefficient for the three judge mean ICC(2,3) is \emph{fair to good} by the criteria of \citet{cicchetti1994} (poor $<0.40$, fair $0.40$--$0.59$, good $0.60$--$0.74$), ranging from $0.48$ to $0.67$ over the 35-model pool ($n = 1{,}023$ artifacts with a complete panel).

\begin{table}[htbp]
\centering
\small
\begin{tabular}{lccc}
\toprule
Dimension & ICC(2,3) & \citet{cicchetti1994} \\
\midrule
Blend semantic integration quality ($Q(c')$)  & 0.67 & good \\
Blend utility ($G(g)$) & 0.67 & good \\
Analogy integration quality ($Q(h)$)  & 0.56 & fair \\
Analogy utility ($V(h)$)  & 0.55 & fair \\
Blend utility ($V(c')$)   & 0.48 & fair \\
\bottomrule
\end{tabular}
\caption{\textbf{Inter-judge reliability of the three judge panel}, as assessed by the intra-class correlation coefficient \citep{cicchetti1994}. ICC(2,3) is the reliability of the mean of the three judges.}
\label{tab:icc}
\end{table}

However, to corroborate this panel against human ratings, an author independently rated a blind subsample of 60 items (30 analogy, 30 blend), applying the same rubric as the judges without access to the model identity or the panel's verdicts. Over all 150 dimension judgments, the human and the panel majority agree \textbf{66\%} of the time. However, on judgments where the panel is unanimous (all three judges agree), human agreement rises to \textbf{75\%} ($n=96$). Restricting the leaderboard to items where the LLM judge panel fully agrees leaves the ordering largely the same, as the rankings on this high-confidence subsample correlate with the full leaderboard at Spearman $\rho=0.945$ and Pearson $r=0.941$ (both $p<10^{-9}$, $n=21$), and even within the top eight the correlation remains strong (Spearman $\rho=0.79$, Pearson $r=0.86$, both $p<0.05$).
\clearpage
\section{Full Leaderboard and Task Scores} \label{app:full_results}
This section reports the full leaderboard and task dimension breakdown, expanding on the top five aggregate task data presented in \Cref{tab:leaderboard}.
% Auto-generated by make_appendix_tables.py -- do not edit by hand.
\providecommand{\provlogo}[1]{\raisebox{-0.15ex}{\includegraphics[height=0.85em]{media/logos/#1}}\,}
\begin{table}[htbp]
\centering
\small
\caption{\textbf{Full leaderboard.} Every model by the per-dimension composite, as a \% of the maximum.}
\label{tab:leaderboard_full}
\resizebox{\ifdim\width>\linewidth\linewidth\else\width\fi}{!}{%
\begin{tabular}{rlcccc}
\toprule
Rank & Model & \textbf{Overall} & Association & Analogy & Blending \\
\midrule
1 & \provlogo{openai.pdf}gpt-5.6-sol & 55.2 & 45.4 & 55.4 & 64.8 \\
2 & \provlogo{openai.pdf}gpt-6-astra & 54.4 & 45.6 & 61.1 & 56.5 \\
3 & \provlogo{claude.pdf}claude-opus-4.6 & 52.0 & 51.4 & 63.1 & 41.4 \\
4 & \provlogo{googlegemini.pdf}gemini-3.1-pro & 48.7 & 39.1 & 47.3 & 59.5 \\
5 & \provlogo{claude.pdf}claude-fable-5 & 48.1 & 47.3 & 51.3 & 45.7 \\
6 & \provlogo{xai.pdf}grok-4.6 & 47.6 & 47.8 & 60.1 & 35.0 \\
7 & \provlogo{openai.pdf}gpt-5 & 47.4 & 46.2 & 58.6 & 37.5 \\
8 & \provlogo{xai.pdf}grok-4.5 & 46.9 & 47.5 & 55.4 & 37.6 \\
9 & \provlogo{googlegemini.pdf}gemini-3.7-flash & 46.4 & 40.9 & 45.2 & 52.9 \\
10 & \provlogo{openai.pdf}gpt-5.2 & 46.1 & 50.3 & 49.0 & 39.1 \\
11 & \provlogo{claude.pdf}claude-opus-4.5 & 44.3 & 48.0 & 47.4 & 37.4 \\
12 & \provlogo{claude.pdf}claude-sonnet-4.6 & 44.1 & 42.6 & 54.3 & 35.5 \\
13 & \provlogo{claude.pdf}claude-fable-5.1 & 42.2 & 44.0 & 39.4 & 43.3 \\
14 & \provlogo{claude.pdf}claude-opus-4.8 & 40.1 & 40.5 & 45.0 & 34.8 \\
15 & \provlogo{claude.pdf}claude-sonnet-4.5 & 39.8 & 44.9 & 36.2 & 38.3 \\
16 & \provlogo{claude.pdf}claude-opus-4.7 & 39.3 & 44.8 & 44.8 & 28.2 \\
17 & \provlogo{claude.pdf}claude-opus-5 & 37.6 & 40.6 & 35.6 & 36.6 \\
18 & \provlogo{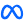}llama-3.3-70b & 36.0 & 45.1 & 38.9 & 24.0 \\
19 & \provlogo{openai.pdf}gpt-5-mini & 35.5 & 47.9 & 38.6 & 19.9 \\
20 & \provlogo{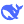}deepseek-r1 & 34.5 & 28.7 & 44.3 & 30.4 \\
21 & \provlogo{claude.pdf}claude-sonnet-5 & 32.7 & 41.7 & 25.7 & 30.8 \\
22 & \provlogo{googlegemini.pdf}gemini-2.5-flash & 31.9 & 48.3 & 28.8 & 18.8 \\
23 & \provlogo{googlegemini.pdf}gemini-3-flash & 30.0 & 31.8 & 36.1 & 22.0 \\
24 & \provlogo{googlegemini.pdf}gemini-2.5-pro & 29.8 & 39.5 & 23.0 & 26.9 \\
25 & \provlogo{openai.pdf}gpt-4.1 & 29.8 & 36.7 & 27.6 & 25.0 \\
26 & \provlogo{deepseek.pdf}deepseek-chat & 29.0 & 31.6 & 27.4 & 28.1 \\
27 & \provlogo{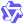}qwen3-max & 27.7 & 30.9 & 32.3 & 19.8 \\
28 & \provlogo{qwen.pdf}qwen-2.5-72b & 27.0 & 31.3 & 27.8 & 22.0 \\
29 & \provlogo{meta.pdf}llama-4-maverick & 26.1 & 36.8 & 22.1 & 19.4 \\
30 & \provlogo{zai.pdf}glm-4.6 & 25.6 & 27.7 & 24.9 & 24.1 \\
31 & \provlogo{openai.pdf}gpt-4o-mini & 24.8 & 32.4 & 30.8 & 11.1 \\
32 & \provlogo{deepseek.pdf}deepseek-v3 & 23.2 & 26.7 & 22.9 & 20.1 \\
33 & \provlogo{zai.pdf}glm-4.5-air & 23.0 & 19.6 & 23.1 & 26.3 \\
34 & \provlogo{kimi.pdf}kimi-k2 & 20.9 & 24.3 & 17.0 & 21.5 \\
35 & \provlogo{microsoft.pdf}phi-4 & 17.4 & 21.9 & 26.0 & 4.4 \\
36 & \textbf{\textit{Humans}} & 13.7 & 14.7 & 19.1 & 7.4 \\
\bottomrule
\end{tabular}
}
\end{table}

\begin{table}[htbp]
\centering
\small
\caption{\textbf{Association full scores.} Every model's score on each dimension, as a \% of the maximum, ordered by the task composite.}
\label{tab:full_association}
\resizebox{\ifdim\width>\linewidth\linewidth\else\width\fi}{!}{%
\begin{tabular}{lcccc}
\toprule
Model & \textbf{Composite} & Utility & Surprise & Originality \\
\midrule
\provlogo{claude.pdf}claude-opus-4.6 & 51.4 & 77.4 & 52.1 & 24.6 \\
\provlogo{openai.pdf}gpt-5.2 & 50.3 & 77.6 & 48.6 & 24.7 \\
\provlogo{googlegemini.pdf}gemini-2.5-flash & 48.3 & 73.0 & 47.4 & 24.4 \\
\provlogo{claude.pdf}claude-opus-4.5 & 48.0 & 72.5 & 49.1 & 22.5 \\
\provlogo{openai.pdf}gpt-5-mini & 47.9 & 74.3 & 46.9 & 22.6 \\
\provlogo{xai.pdf}grok-4.6 & 47.8 & 71.8 & 48.5 & 23.0 \\
\provlogo{xai.pdf}grok-4.5 & 47.5 & 71.4 & 47.2 & 23.9 \\
\provlogo{claude.pdf}claude-fable-5 & 47.3 & 70.3 & 47.0 & 24.6 \\
\provlogo{openai.pdf}gpt-5 & 46.2 & 70.2 & 45.7 & 22.7 \\
\provlogo{openai.pdf}gpt-6-astra & 45.6 & 67.4 & 44.7 & 24.7 \\
\provlogo{openai.pdf}gpt-5.6-sol & 45.4 & 69.0 & 43.7 & 23.4 \\
\provlogo{meta.pdf}llama-3.3-70b & 45.1 & 70.1 & 43.9 & 21.3 \\
\provlogo{claude.pdf}claude-sonnet-4.5 & 44.9 & 67.4 & 46.0 & 21.4 \\
\provlogo{claude.pdf}claude-opus-4.7 & 44.8 & 67.5 & 46.2 & 20.8 \\
\provlogo{claude.pdf}claude-fable-5.1 & 44.0 & 65.4 & 43.7 & 22.8 \\
\provlogo{claude.pdf}claude-sonnet-4.6 & 42.6 & 63.9 & 43.2 & 20.7 \\
\provlogo{claude.pdf}claude-sonnet-5 & 41.7 & 63.5 & 42.2 & 19.3 \\
\provlogo{googlegemini.pdf}gemini-3.7-flash & 40.9 & 60.4 & 41.6 & 20.8 \\
\provlogo{claude.pdf}claude-opus-5 & 40.6 & 60.3 & 40.2 & 21.4 \\
\provlogo{claude.pdf}claude-opus-4.8 & 40.5 & 63.0 & 40.2 & 18.4 \\
\provlogo{googlegemini.pdf}gemini-2.5-pro & 39.5 & 57.8 & 39.3 & 21.5 \\
\provlogo{googlegemini.pdf}gemini-3.1-pro & 39.1 & 56.9 & 38.6 & 21.9 \\
\provlogo{meta.pdf}llama-4-maverick & 36.8 & 56.5 & 36.8 & 17.1 \\
\provlogo{openai.pdf}gpt-4.1 & 36.7 & 55.1 & 37.3 & 17.5 \\
\provlogo{openai.pdf}gpt-4o-mini & 32.4 & 49.6 & 31.9 & 15.6 \\
\provlogo{googlegemini.pdf}gemini-3-flash & 31.8 & 46.6 & 32.6 & 16.2 \\
\provlogo{deepseek.pdf}deepseek-chat & 31.6 & 47.9 & 32.0 & 15.0 \\
\provlogo{qwen.pdf}qwen-2.5-72b & 31.3 & 47.2 & 32.3 & 14.4 \\
\provlogo{qwen.pdf}qwen3-max & 30.9 & 45.3 & 32.2 & 15.4 \\
\provlogo{deepseek.pdf}deepseek-r1 & 28.7 & 43.2 & 28.4 & 14.4 \\
\provlogo{zai.pdf}glm-4.6 & 27.7 & 41.8 & 28.4 & 12.9 \\
\provlogo{deepseek.pdf}deepseek-v3 & 26.7 & 40.3 & 27.7 & 12.2 \\
\provlogo{kimi.pdf}kimi-k2 & 24.3 & 35.6 & 25.0 & 12.3 \\
\provlogo{microsoft.pdf}phi-4 & 21.9 & 33.5 & 21.7 & 10.4 \\
\provlogo{zai.pdf}glm-4.5-air & 19.6 & 29.8 & 19.3 & 9.7 \\
\textbf{\textit{Humans}} & 14.7 & 22.1 & 14.0 & 7.8 \\
\bottomrule
\end{tabular}
}
\end{table}

\begin{table}[htbp]
\centering
\small
\caption{\textbf{Analogy full scores.} Every model's score on each dimension, as a \% of the maximum, ordered by the task composite.}
\label{tab:full_analogy}
\resizebox{\ifdim\width>\linewidth\linewidth\else\width\fi}{!}{%
\begin{tabular}{lccccccc}
\toprule
Model & \textbf{Composite} & Utility & Surprise & Orig.\ (base) & Orig.\ (emrg.) & Emerg.\ integ. & Emerg.\ util. \\
\midrule
\provlogo{claude.pdf}claude-opus-4.6 & 63.1 & 86.7 & 65.1 & 39.3 & 44.0 & 63.3 & 80.0 \\
\provlogo{openai.pdf}gpt-6-astra & 61.1 & 76.7 & 61.2 & 39.7 & 42.5 & 70.0 & 76.7 \\
\provlogo{xai.pdf}grok-4.6 & 60.1 & 76.7 & 59.6 & 36.3 & 41.1 & 76.7 & 70.0 \\
\provlogo{openai.pdf}gpt-5 & 58.6 & 76.7 & 60.9 & 37.8 & 39.5 & 70.0 & 66.7 \\
\provlogo{xai.pdf}grok-4.5 & 55.4 & 70.0 & 53.8 & 33.1 & 35.8 & 70.0 & 70.0 \\
\provlogo{openai.pdf}gpt-5.6-sol & 55.4 & 70.0 & 53.8 & 33.0 & 35.8 & 70.0 & 70.0 \\
\provlogo{claude.pdf}claude-sonnet-4.6 & 54.3 & 70.0 & 56.6 & 32.5 & 36.8 & 60.0 & 70.0 \\
\provlogo{claude.pdf}claude-fable-5 & 51.3 & 70.0 & 54.0 & 31.9 & 35.3 & 53.3 & 63.3 \\
\provlogo{openai.pdf}gpt-5.2 & 49.0 & 63.3 & 49.6 & 29.8 & 31.3 & 60.0 & 60.0 \\
\provlogo{claude.pdf}claude-opus-4.5 & 47.4 & 63.3 & 46.0 & 25.9 & 32.2 & 53.3 & 63.3 \\
\provlogo{googlegemini.pdf}gemini-3.1-pro & 47.3 & 60.0 & 46.2 & 29.4 & 31.8 & 60.0 & 56.7 \\
\provlogo{googlegemini.pdf}gemini-3.7-flash & 45.2 & 60.0 & 46.2 & 27.8 & 30.5 & 50.0 & 56.7 \\
\provlogo{claude.pdf}claude-opus-4.8 & 45.0 & 63.3 & 46.6 & 26.7 & 29.9 & 53.3 & 50.0 \\
\provlogo{claude.pdf}claude-opus-4.7 & 44.8 & 63.3 & 47.5 & 26.8 & 31.4 & 46.7 & 53.3 \\
\provlogo{deepseek.pdf}deepseek-r1 & 44.3 & 60.0 & 48.6 & 29.8 & 30.7 & 36.7 & 60.0 \\
\provlogo{claude.pdf}claude-fable-5.1 & 39.4 & 50.0 & 39.1 & 24.3 & 26.4 & 46.7 & 50.0 \\
\provlogo{meta.pdf}llama-3.3-70b & 38.9 & 56.7 & 39.3 & 22.8 & 27.7 & 30.0 & 56.7 \\
\provlogo{openai.pdf}gpt-5-mini & 38.6 & 53.3 & 39.1 & 23.5 & 25.8 & 43.3 & 46.7 \\
\provlogo{claude.pdf}claude-sonnet-4.5 & 36.2 & 50.0 & 37.6 & 24.3 & 25.2 & 30.0 & 50.0 \\
\provlogo{googlegemini.pdf}gemini-3-flash & 36.1 & 50.0 & 39.1 & 24.3 & 26.5 & 33.3 & 43.3 \\
\provlogo{claude.pdf}claude-opus-5 & 35.6 & 53.3 & 41.7 & 26.6 & 28.7 & 30.0 & 33.3 \\
\provlogo{qwen.pdf}qwen3-max & 32.3 & 46.7 & 36.8 & 22.4 & 24.6 & 23.3 & 40.0 \\
\provlogo{openai.pdf}gpt-4o-mini & 30.8 & 46.7 & 34.1 & 19.0 & 21.8 & 20.0 & 43.3 \\
\provlogo{googlegemini.pdf}gemini-2.5-flash & 28.8 & 40.0 & 30.9 & 18.3 & 20.2 & 26.7 & 36.7 \\
\provlogo{qwen.pdf}qwen-2.5-72b & 27.8 & 40.0 & 29.8 & 17.4 & 19.5 & 23.3 & 36.7 \\
\provlogo{openai.pdf}gpt-4.1 & 27.6 & 40.0 & 30.2 & 19.4 & 19.2 & 16.7 & 40.0 \\
\provlogo{deepseek.pdf}deepseek-chat & 27.4 & 40.0 & 28.4 & 16.4 & 16.1 & 23.3 & 40.0 \\
\provlogo{microsoft.pdf}phi-4 & 26.0 & 40.0 & 29.9 & 16.7 & 19.1 & 10.0 & 40.0 \\
\provlogo{claude.pdf}claude-sonnet-5 & 25.7 & 36.7 & 28.4 & 17.4 & 18.6 & 23.3 & 30.0 \\
\provlogo{zai.pdf}glm-4.6 & 24.9 & 36.7 & 28.4 & 17.8 & 20.0 & 13.3 & 33.3 \\
\provlogo{zai.pdf}glm-4.5-air & 23.1 & 36.7 & 26.3 & 17.1 & 18.3 & 3.3 & 36.7 \\
\provlogo{googlegemini.pdf}gemini-2.5-pro & 23.0 & 30.0 & 22.7 & 15.3 & 16.4 & 26.7 & 26.7 \\
\provlogo{deepseek.pdf}deepseek-v3 & 22.9 & 36.7 & 28.1 & 14.9 & 17.6 & 20.0 & 20.0 \\
\provlogo{meta.pdf}llama-4-maverick & 22.1 & 30.0 & 21.8 & 13.2 & 14.1 & 23.3 & 30.0 \\
\textbf{\textit{Humans}} & 19.1 & 32.9 & 23.1 & 15.0 & 18.8 & 5.4 & 19.4 \\
\provlogo{kimi.pdf}kimi-k2 & 17.0 & 26.7 & 20.7 & 13.7 & 14.5 & 10.0 & 16.7 \\
\bottomrule
\end{tabular}
}
\end{table}

\begin{table}[htbp]
\centering
\small
\caption{\textbf{Blending full scores.} Every model's score on each dimension, as a \% of the maximum, ordered by the task composite.}
\label{tab:full_blending}
\resizebox{\ifdim\width>\linewidth\linewidth\else\width\fi}{!}{%
\begin{tabular}{lccccccc}
\toprule
Model & \textbf{Composite} & Utility & Surprise & Orig.\ (base) & Orig.\ (emrg.) & Emerg.\ integ. & Emerg.\ util. \\
\midrule
\provlogo{openai.pdf}gpt-5.6-sol & 64.8 & 83.3 & 70.5 & 35.7 & 42.7 & 76.7 & 80.0 \\
\provlogo{googlegemini.pdf}gemini-3.1-pro & 59.5 & 83.3 & 66.6 & 30.4 & 38.5 & 55.0 & 83.3 \\
\provlogo{openai.pdf}gpt-6-astra & 56.5 & 73.3 & 60.5 & 31.7 & 38.4 & 65.0 & 70.0 \\
\provlogo{googlegemini.pdf}gemini-3.7-flash & 52.9 & 70.0 & 57.7 & 26.7 & 33.3 & 60.0 & 70.0 \\
\provlogo{claude.pdf}claude-fable-5 & 45.7 & 60.0 & 48.1 & 21.3 & 28.2 & 56.7 & 60.0 \\
\provlogo{claude.pdf}claude-fable-5.1 & 43.3 & 56.7 & 44.7 & 23.5 & 26.8 & 51.7 & 56.7 \\
\provlogo{claude.pdf}claude-opus-4.6 & 41.4 & 56.7 & 45.3 & 20.6 & 27.6 & 41.7 & 56.7 \\
\provlogo{openai.pdf}gpt-5.2 & 39.1 & 53.3 & 41.5 & 19.9 & 24.9 & 41.7 & 53.3 \\
\provlogo{claude.pdf}claude-sonnet-4.5 & 38.3 & 53.3 & 43.4 & 18.9 & 24.4 & 40.0 & 50.0 \\
\provlogo{xai.pdf}grok-4.5 & 37.6 & 50.0 & 40.6 & 21.2 & 24.1 & 40.0 & 50.0 \\
\provlogo{openai.pdf}gpt-5 & 37.5 & 50.0 & 40.4 & 19.4 & 25.2 & 43.3 & 46.7 \\
\provlogo{claude.pdf}claude-opus-4.5 & 37.4 & 50.0 & 39.8 & 17.6 & 23.5 & 43.3 & 50.0 \\
\provlogo{claude.pdf}claude-opus-5 & 36.6 & 50.0 & 42.5 & 20.8 & 26.1 & 33.3 & 46.7 \\
\provlogo{claude.pdf}claude-sonnet-4.6 & 35.5 & 46.7 & 37.6 & 17.9 & 22.6 & 41.7 & 46.7 \\
\provlogo{xai.pdf}grok-4.6 & 35.0 & 46.7 & 37.2 & 21.4 & 23.3 & 38.3 & 43.3 \\
\provlogo{claude.pdf}claude-opus-4.8 & 34.8 & 46.7 & 37.1 & 15.4 & 21.4 & 41.7 & 46.7 \\
\provlogo{claude.pdf}claude-sonnet-5 & 30.8 & 40.0 & 31.8 & 13.8 & 19.1 & 40.0 & 40.0 \\
\provlogo{deepseek.pdf}deepseek-r1 & 30.4 & 43.3 & 34.8 & 17.9 & 19.8 & 26.7 & 40.0 \\
\provlogo{claude.pdf}claude-opus-4.7 & 28.2 & 36.7 & 28.8 & 12.0 & 18.6 & 36.7 & 36.7 \\
\provlogo{deepseek.pdf}deepseek-chat & 28.1 & 40.0 & 31.0 & 13.0 & 17.6 & 26.7 & 40.0 \\
\provlogo{googlegemini.pdf}gemini-2.5-pro & 26.9 & 40.0 & 32.5 & 16.8 & 20.3 & 15.0 & 36.7 \\
\provlogo{zai.pdf}glm-4.5-air & 26.3 & 36.7 & 29.0 & 15.6 & 16.5 & 26.7 & 33.3 \\
\provlogo{openai.pdf}gpt-4.1 & 25.0 & 33.3 & 26.3 & 12.2 & 16.7 & 28.3 & 33.3 \\
\provlogo{zai.pdf}glm-4.6 & 24.1 & 33.3 & 27.6 & 12.9 & 15.7 & 25.0 & 30.0 \\
\provlogo{meta.pdf}llama-3.3-70b & 24.0 & 33.3 & 25.7 & 13.5 & 16.6 & 21.7 & 33.3 \\
\provlogo{qwen.pdf}qwen-2.5-72b & 22.0 & 33.3 & 24.0 & 11.8 & 16.1 & 16.7 & 30.0 \\
\provlogo{googlegemini.pdf}gemini-3-flash & 22.0 & 30.0 & 25.7 & 11.0 & 15.1 & 23.3 & 26.7 \\
\provlogo{kimi.pdf}kimi-k2 & 21.5 & 30.0 & 24.7 & 11.8 & 14.0 & 18.3 & 30.0 \\
\provlogo{deepseek.pdf}deepseek-v3 & 20.1 & 26.7 & 19.5 & 8.6 & 12.4 & 26.7 & 26.7 \\
\provlogo{openai.pdf}gpt-5-mini & 19.9 & 26.7 & 21.4 & 8.8 & 12.8 & 23.3 & 26.7 \\
\provlogo{qwen.pdf}qwen3-max & 19.8 & 26.7 & 21.6 & 10.2 & 11.8 & 21.7 & 26.7 \\
\provlogo{meta.pdf}llama-4-maverick & 19.4 & 30.0 & 22.3 & 11.0 & 14.7 & 11.7 & 26.7 \\
\provlogo{googlegemini.pdf}gemini-2.5-flash & 18.8 & 26.7 & 21.4 & 10.1 & 12.8 & 15.0 & 26.7 \\
\provlogo{openai.pdf}gpt-4o-mini & 11.1 & 16.7 & 11.5 & 5.8 & 7.7 & 8.3 & 16.7 \\
\textbf{\textit{Humans}} & 7.4 & 12.8 & 9.7 & 6.7 & 6.3 & 1.3 & 7.7 \\
\provlogo{microsoft.pdf}phi-4 & 4.4 & 6.7 & 4.9 & 2.2 & 2.8 & 3.3 & 6.7 \\
\bottomrule
\end{tabular}
}
\end{table}

\clearpage
\section{Robustness to the Choice of Embedding Model} \label{app:embedding_robustness}
Here, we test the robustness of surprise and originality scores to varying embedding models. To test this, we re-scored surprise, originality, and emergent originality for every artifact under two encoders from different model families: \texttt{bge-small-en-v1.5} \citep{bge_embedding} and \texttt{multilingual-e5-small} \citep{wang2024multilingual}. As shown in \Cref{tab:encoder_leaderboard}, the overall ranking of the 35 models is nearly identical under all three encoders (Spearman $\rho \geq 0.998$).

\begin{table}[H]
\centering
\small
\begin{tabular}{lcccc}
\toprule
& Overall & \multicolumn{3}{c}{Per-task} \\
\cmidrule(lr){2-2} \cmidrule(lr){3-5}
Encoder pair & $\rho$ & Assoc. & Analogy & Blend \\
\midrule
MiniLM $\sim$ BGE & 0.999 & 0.996 & 0.997 & 0.999 \\
MiniLM $\sim$ E5 & 0.998 & 0.996 & 0.996 & 0.997 \\
BGE $\sim$ E5 & 0.998 & 0.996 & 0.999 & 0.999 \\
\bottomrule
\end{tabular}
\caption{\textbf{Leaderboard agreement across embedding models.} Spearman $\rho$ between the overall rankings of the 35 models under each pair of encoders, as well as the Spearman $\rho$ by task.}
\label{tab:encoder_leaderboard}
\end{table}

\paragraph{Robustness of association originality predictive power results to embedding choice} We also tested the robustness of the finding that association originality predicts every other facet (\Cref{sec:results}), including a ``crossed'' version in which association originality is scored under one encoder and the other facets under another, to rule out shared embedding geometry as an explanation. \Cref{tab:encoder_facets} reports the correlations with association originality scored under \texttt{all-MiniLM-L6-v2} and the remaining facets scored under all of the three total encoders (including \texttt{all-MiniLM-L6-v2}). Every novelty facet of analogy and blending, as well as blend utility, remains significantly correlated with association originality under every encoder, whereas analogy utility remains uncorrelated ($r \leq 0.22$). The mean cross-task correlation ranges from $\bar{r} = 0.41$ to $0.48$ over all nine encoder pairs.

\begin{table}[H]
\centering
\small
\begin{tabular}{lccc}
\toprule
& \multicolumn{3}{c}{Facet scored under} \\
\cmidrule(lr){2-4}
Facet & MiniLM & BGE & E5 \\
\midrule
Analogy surprise & 0.62$^{***}$ & 0.39$^{*}$ & 0.42$^{*}$ \\
Analogy originality & 0.72$^{***}$ & 0.65$^{***}$ & 0.60$^{***}$ \\
Analogy emergent originality & 0.70$^{***}$ & 0.66$^{***}$ & 0.69$^{***}$ \\
Blend utility & 0.48$^{**}$ & 0.48$^{**}$ & 0.48$^{**}$ \\
Blend surprise & 0.50$^{**}$ & 0.51$^{**}$ & 0.45$^{**}$ \\
Blend originality & 0.39$^{*}$ & 0.46$^{**}$ & 0.37$^{*}$ \\
Blend emergent originality & 0.36$^{*}$ & 0.35$^{*}$ & 0.43$^{**}$ \\
\midrule
Mean cross-task $\bar{r}$ & 0.47 & 0.44 & 0.41 \\
\bottomrule
\end{tabular}
\caption{\textbf{Association originality predicts the other facets under every encoder.} Pearson $r$ across the 35 models between association originality scored under \texttt{all-MiniLM-L6-v2} and each facet scored under the column's encoder, and $\bar{r}$ averages the six base facets of the other two tasks. All correlations are statistically significant, with significance levels denoted by: $^{*}p<0.05$, $^{**}p<0.01$, $^{***}p<0.001$.}
\label{tab:encoder_facets}
\end{table}

\end{document}